\documentclass[10pt,twocolumn]{article}

\usepackage{times}
\usepackage{epsfig}
\usepackage{graphicx}
\usepackage{amsmath}
\usepackage{amssymb}

\usepackage{caption}
\usepackage[binary-units,per-mode=symbol]{siunitx}
\usepackage[modulo]{lineno}
\usepackage{nicefrac}
\usepackage{textcomp}
\usepackage{url}
\usepackage[dvipsnames]{xcolor}
\usepackage{geometry}
\usepackage[pagebackref=true,breaklinks=true,colorlinks,allcolors=cyan,bookmarks=false]{hyperref}
\newcommand*{\nolink}[1]{%
  {\protect\NoHyper#1\protect\endNoHyper}%
}

\global\hyphenpenalty=100
\renewcommand*{\thefootnote}{\fnsymbol{footnote}}
\newcommand{\taud}{\tau_\text{d}}

\newcommand{\atul}[1]{\textcolor{black}{#1}}

\begin{document}

\newcommand{\mytitle}{Passive Inter-Photon Imaging}

\title{\mytitle}

\author{
  Atul Ingle$^{\,\text{a}\,\ast}$, Trevor Seets$\,^\text{a}$, Mauro Buttafava$\,^\text{b}$,
  Shantanu Gupta$\,^\text{a}$,\\
  Alberto Tosi$\,^\text{b}$, Mohit Gupta$\,^{\text{a}\,\dagger}$,
  Andreas Velten$\,^{\text{a}\,\dagger}$ \vspace{2pt} \\
  {\normalsize $^\text{a}$ University of Wisconsin-Madison} \,\,\,\,
  {\normalsize $^\text{b}$ Politecnico di Milano}
}
\date{}
\maketitle
\renewcommand*{\thefootnote}{$\dagger$}
\setcounter{footnote}{1}
\footnotetext{Equal contribution. \newline	
	This research was supported in part by DARPA HR0011-16-C-0025, DoE NNSA
	DE-NA0003921 \cite{NNSADisclaimer}, NSF GRFP DGE-1747503, NSF CAREER
	1846884 and 1943149, and Wisconsin Alumni Research Foundation. \newline
  $^\ast$ Email: \texttt{ingle@uwalumni.com}
}

\renewcommand*{\thefootnote}{\arabic{footnote}}
\setcounter{footnote}{0}

\newenvironment{myitemize}
{ \begin{itemize}
    \setlength{\itemsep}{3pt}
    \setlength{\parskip}{2pt}
    \setlength{\parsep}{3pt}     }
{ \end{itemize}                  }

\begin{abstract}

Digital camera pixels measure image intensities by converting incident
light energy into an analog electrical current, and then digitizing it
into a fixed-width binary representation.  This direct measurement method,
while conceptually simple, suffers from limited dynamic range and poor
performance under extreme illumination --- electronic noise dominates under
low illumination, and pixel full-well capacity results in saturation
under bright illumination. We propose a novel intensity cue based on
measuring \textnormal{inter-photon~timing}, defined as the time delay between
detection of successive photons. Based on the statistics of inter-photon
times measured by a time-resolved single-photon sensor, we develop theory and
algorithms for a scene brightness estimator which works over extreme dynamic
range; we experimentally demonstrate imaging scenes with a dynamic range of
over ten million to one. The proposed techniques, aided by the emergence of
single-photon sensors such as single-photon avalanche diodes (SPADs) with
picosecond timing resolution, will have implications for a wide range of
imaging applications: robotics, consumer photography, astronomy, microscopy
and biomedical imaging.

\end{abstract}



\section{Measuring Light from Darkness}
\begin{figure*}[!ht]
\includegraphics[width=0.98\textwidth]{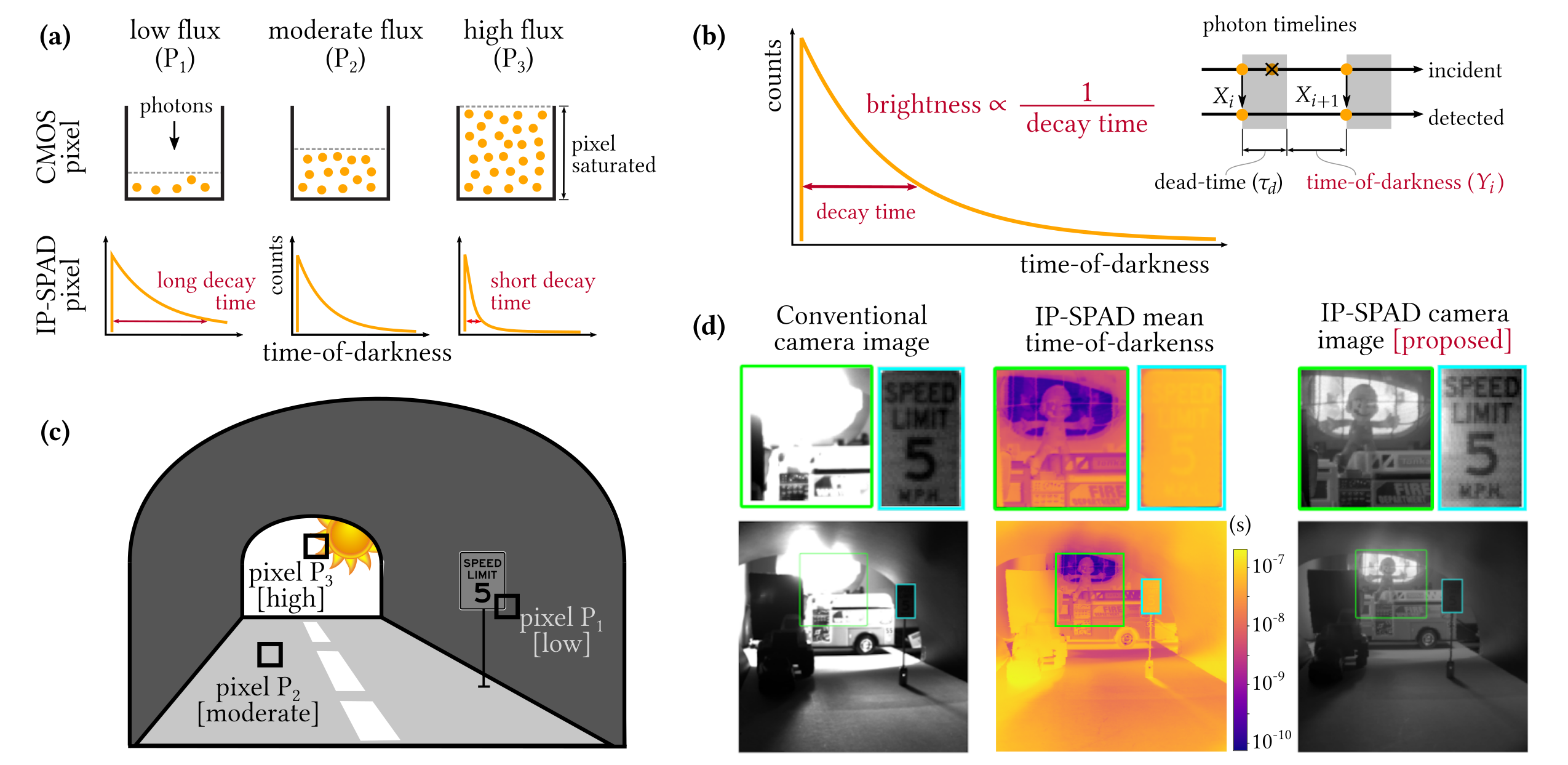}
\caption{\textbf{Passive imaging with an inter-photon single-photon
avalanche diode (IP-SPAD):} (a) A conventional image
sensor pixel estimates scene brightness using a well-filling mechanism;
the well has a finite capacity and saturates for very high brightness
levels. (b) An IP-SPAD measures scene brightness from inter-photon timing
measurements that follow Poisson statistics. The higher the brightness, the
smaller the inter-photon time, the faster the decay rate of the inter-photon
histogram. By capturing photon timing information with very high precision,
this estimator can provide scene brightness estimates well
beyond the saturation limit of conventional pixels.  (c) A representative
extreme dynamic range scene of a tunnel with three different flux levels
(low, moderate and high) shown for illustration.  (d) Experimental results
from our hardware prototype comparing a conventional CMOS camera image and an
image obtained from our IP-SPAD prototype.
\label{fig:teaser}
\vspace{-0.15in} }
\end{figure*}

Digital camera technology has witnessed a remarkable revolution in terms
of size, cost and image quality over the past few years. Throughout this
progress, however, one fundamental characteristic of a camera sensor
has not changed: the way a camera pixel measures brightness.
Conventional image sensor pixels manufactured with complementary metal oxide
semiconductor (CMOS) and charge-coupled device (CCD) technology can be thought
of as light buckets (Fig.~\ref{fig:teaser}(a)), which measure scene brightness
in two steps: first, they collect hundreds or thousands of photons and convert
the energy into an analog electrical signal (e.g.  current or voltage), and
then they digitize this analog quantity using high-resolution analog-to-digital
converters. Conceptually, there are two main drawbacks of this image formation
strategy. First, at extremely low brightness levels, noise in the pixel
electronics dominates resulting in poor signal-to-noise-ratio (SNR). Second,
since each pixel bucket has a fixed maximum capacity, bright regions in the
scene cause the pixels to saturate and subsequent incident photons do not get
recorded.  

In this paper, we explore a different approach for measuring image intensities.
Instead of estimating intensities directly from the number of photons incident
on a pixel, we propose a novel intensity cue based on \emph{inter-photon
timing}, defined as the time delay between detection of successive photons.
Intuitively, as the brightness increases, the \emph{time-of-darkness} between
consecutive photon detections decreases.  By modeling the statistics of photon
arrivals, we derive a theoretical expressions that relates the average
inter-photon delay and the incident flux. The key observation is that because
photon arrivals are stochastic, the average inter-photon time
decreases asymptotically as the incident flux increases.
Using this novel temporal intensity cue, we design algorithms to estimate
brightness from as few as one photon timestamp per pixel to extremely high
brightness, beyond the saturation limit of conventional sensors.

\smallskip
\noindent {\bf How to Measure Inter-Photon Timing?} The inter-photon
timing intensity cue and the resulting brightness estimators can achieve
extremely high dynamic range. A natural question to ask then is: How does one
measure the inter-photon timing? Conventional CMOS sensor pixels do not have
the ability to measure time delays between individual photons at the timing
granularity needed for estimating intensities with high precision. Fortunately,
there is an emerging class of sensors called single-photon avalanche diodes
(SPADs) \cite{cova1996avalanche,bruschini2019single}, that can not only detect
individual photons, but also precisely time-tag each captured photon with
picosecond resolution.



\smallskip \noindent {\bf Emergence of Single-Photon Sensors:} SPADs are
naturally suited for imaging in low illumination conditions, and thus, are fast
becoming the sensors of choice for applications that require extreme
sensitivity to photons together with fine-grained temporal information:
\atul{single-photon 3D time-of-flight imaging
\cite{yoshida_2020,lindner2018252,tachella2019real,rapp2020advances,lindell2020three},}
transient imaging \cite{Ulku_2019,turpin2020spatial}, non-line-of-sight imaging
\cite{liu2019non,grau2020deep}, and fluorescence microscopy
\cite{perenzoni2015160}. While these applications use SPADs in active imaging
setups in synchronization with an illumination source such as a pulsed
laser, recently these sensors have been explored as passive, general-purpose
imaging devices for high-speed and high-dynamic range photography
\cite{Antolovic_2018,ingle2019high,ma2020quanta}.  In particular, it was shown
that SPADs can be used to measure incident flux while operating as passive,
free-running pixels (PF-SPAD imaging) \cite{ingle2019high}. The dynamic range
of the resulting measurements, although higher than conventional pixels (that
rely on a finite-depth well filling light detection method like CCD and CMOS
sensors), is inherently limited due to \emph{quantization} stemming from the
discrete nature of photon counts.

\smallskip
\noindent {\bf Intensity from Inter-Photon Timing:} Our key idea is that
it is possible to circumvent the limitations of counts-based photon flux
estimation by exploiting photon timing information from a SPAD. The
additional time-dimension is a rich source of information that is
inaccessible to conventional photon count-based methods. We derive a scene
brightness estimator that relies on the decay time statistics of the
inter-photon times captured by a SPAD sensor as shown in
Fig.~\ref{fig:teaser}(b).  We call our image sensing method
\emph{inter-photon SPAD (IP-SPAD)} imaging. An IP-SPAD pixel captures the
decay time distribution which gets narrower with increasing brightness. As
shown in Fig.~\ref{fig:teaser}(d), the measurements can be summarized in
terms of the mean time-of-darkness, which can then be used to estimate
incident flux. 

Unlike a photon-counting PF-SPAD pixel whose measurements are inherently
discrete, an IP-SPAD measures decay times as floating point values, capturing
information at much finer granularity than integer-valued counts, thus enabling
measurement of extremely high flux values. In practice, the dynamic range of an
IP-SPAD is limited only by the precision of the floating point representation
used for measuring the time-of-darkness between consecutive photons. Coupled
with the sensitivity of SPADs to individual photons and lower noise compared to
conventional sensors, the proposed approach, for the first time, achieves ultra
high-dynamic range. We experimentally demonstrate a dynamic range of over ten
million to one, simultaneously imaging extremely dark (pixels
P\textsubscript{1} and P\textsubscript{2} inside the tunnel in
Fig.~\ref{fig:teaser}(c)) as well as very bright scene regions (pixel
P\textsubscript{3} outside the tunnel in Fig.~\ref{fig:teaser}(c)).

\section{Related Work}
\noindent\textbf{High-Dynamic-Range Imaging:}
Conventional high-dynamic-range (HDR) imaging techniques using CMOS image
sensors use variable exposures to capture scenes with extreme dynamic range.
The most common method called exposure bracketing
\cite{Gupta_2013,Hasinoff2010} captures multiple images with different exposure
times; shorter exposures reliably capture bright pixels in the scene avoiding
saturation, while longer exposures capture darker pixels while avoiding photon
noise. Another technique involves use of neutral density (ND) filters of
varying densities resulting in a tradeoff between spatial resolution and
dynamic range \cite{Nayar_2000}. ND filters reduce overall sensitivity to avoid
saturation, at the cost of making darker scene pixels noisier.  In contrast, an
IP-SPAD captures scene intensities in a different way by relying on the
non-linear reciprocal relationship between inter-photon timing and scene
brightness. This gives extreme dynamic range in a single capture.

\smallskip
\noindent\textbf{Passive Imaging with Photon-Counting Sensors:}
Previous work on passive imaging with photon counting sensors relies on two
sensor technologies---SPADs and quanta-image sensors (QISs)
\cite{gnanasambandam2019megapixel}. A QIS has single-photon sensitivity but
much lower time resolution than a SPAD pixel. On the other hand, QIS pixels can
be designed with much smaller pixel pitch compared to SPAD pixels, allowing
spatial averaging to further improve dynamic range while still maintaining high
spatial resolution \cite{ma2020quanta}. SPAD-based high-dynamic range schemes
provide lower spatial resolution than the QIS-based counterparts
\cite{dutton2018high}, although, recently, megapixel SPAD arrays capable of
passive photon counting have also be developed \cite{Morimoto_2020}. Previous
work \cite{ingle2019high} has shown that passive free-running SPADs can
potentially provide several orders of magnitude improved dynamic range compared
to conventional CMOS image sensor pixel. The present work exploits the precise
\emph{timing information}, in addition to photon counts, measured by a
free-running SPAD sensor. An IP-SPAD can image scenes with even higher dynamic
range than the counts-based PF-SPAD method.

\smallskip
\noindent\textbf{Methods Relying on Photon Timing:}
The idea of using timing information for passive imaging has been explored
before for conventional CMOS image sensor pixels. A saturated CMOS pixel's
output is simply a constant and meaningless, but if the time taken to reach
saturation is also available \cite{culurciello2003biomorphic}, it provides
information about scene brightness, because a brighter scene point will reach
saturation more quickly (on average) than a dimmer scene point. The idea of
using photon timing information for HDR has also been discussed before but the
dynamic range improvements were limited by the low timing resolution of the
pixels \cite{zarghami2019high, laurenzis2019single} at which point, the photon
timing provides no additional information over photon counts.
\atul{In their seminal work on single-photon 3D imaging
under high illumination conditions,
Rapp \emph{et al.} provide rigorous theoretical derivations of photon-timing-based
maximum-likelihood flux estimators that include
dead-time effects of both the SPAD pixel and the
timing electronics \cite{rapp2019dead,rapp2021high}.
Our theoretical SNR analysis and experimental results show that such photon-timing-based flux estimators can provide extreme dynamic range for passive 2D intensity imaging.}

\smallskip
\noindent\textbf{Methods Relying on Non-linear Sensor Response:}
Logarithmic image sensors include additional pixel electronics that apply
log-compression to capture a large dynamic range. These pixels are difficult to
calibrate and require additional pixel electronics compared to conventional
CMOS image sensor pixels \cite{kavadias2000logarithmic}. A modulo-camera
\cite{zhao2015unbounded} allows a conventional CMOS pixel output to wrap around
after saturation. It requires additional in-pixel computation involving an
iterative algorithm that unwraps the modulo-compression to reconstruct the
high-dynamic-range scene. In contrast, our timing-based HDR flux estimator is a
closed-form expression that can be computed using simple arithmetic operations.
Although our method also requires additional in-pixel electronics to capture
high-resolution timing information, recent trends in SPAD technology indicate
that such arrays can be manufactured cheaply and at scale using CMOS
fabrication techniques \cite{Henderson_2019, Henderson_2019_ISSCC}.

\smallskip
\noindent\textbf{Active Imaging Methods:}
Photon timing information captured by a SPAD sensor has been exploited for
various active imaging applications like transient imaging
\cite{o2017reconstructing}, fluorescence lifetime microscopy
\cite{bruschini2019single}, 3D imaging LiDAR \cite{Kirmani_2013,
gupta2019asynchronous} and non-line-of-sight imaging \cite{kirmani2009looking,
buttafava2015non}.  Active methods capture photon timing information with
respect to a synchronized light source like a pulsed laser that illuminates the
scene. \atul{In a low
flux setting, photon timing information measured with respect to a
synchronized laser source can be used to reconstruct scene depth and reflectivity maps
\cite{Kirmani_2013}. In contrast, here we show that inter-photon timing
information captured with an asynchronous SPAD sensor can enable scene
intensity estimation not just in low but also extremely high flux levels in a
passive imaging setting.}

\begin{figure*}[!t]
  \centering \includegraphics[width=1.00\linewidth]{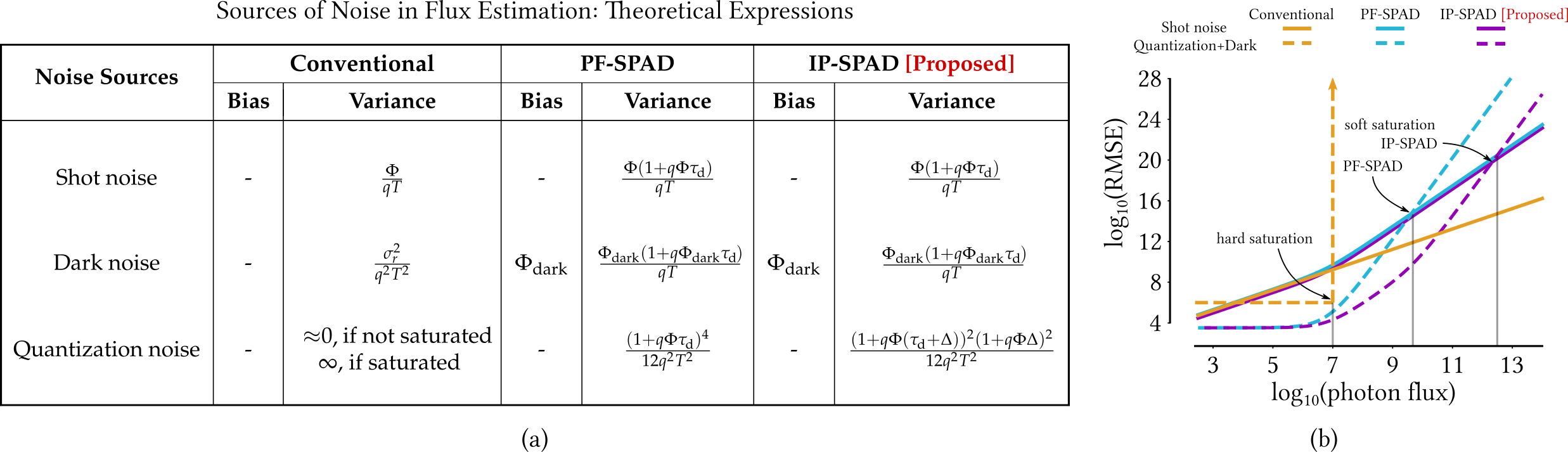}
  \caption{\textbf{Comparison of noise sources in different image sensor pixels:}
	(a) Theoretical expressions for the three main sources of noise affecting a
conventional pixel, PF-SPAD pixel \cite{ingle2019high} and the proposed IP-SPAD
pixel are summarized in this table. Note that the IP-SPAD's sources of noise are similar
to a PF-SPAD except for quantization noise. (b) The expressions in (a) are plotted for
the case of $T=\SI{5}{\ms}$, $q=100\%$, $\sigma_r=5 e^-$, $\Phi_\text{dark}=10$ photons/second,
$\taud = \SI{150}{\ns}$, $\Delta=\SI{200}{\ps}$. The conventional sensor's saturation capacity
is set at 34,000 $e^-$ which matches the maximum possible SPAD counts of
$\lceil \nicefrac{T}{\taud} \rceil$.
Observe that the IP-SPAD soft-saturation point is at a much higher flux level
than the PF-SPAD. \label{fig:sources_of_noise}}
\end{figure*}

\section{Image Formation with Inter-Photon Timing}
\subsection{Flux Estimator \label{sec:flux_estimator}}
Consider a single IP-SPAD pixel passively capturing photons over a fixed
exposure time $T$ from a scene point\footnote{We assume that there is no scene
or camera motion so that the flux $\Phi$ stays constant over the exposure time
$T$.} with true photon flux of $\Phi$ photons per second.  After each photon
detection event, the IP-SPAD pixel goes blind for a fixed duration $\taud$
called the dead-time. During this dead-time, the pixel is reset and the pixel's
time-to-digital converter (TDC)
circuit stores a picosecond resolution timestamp of the most recent
photon detection time, and also increments the total photon count. This process
is repeated until the end of the exposure time $T$. Let $N_T \geq 2$ denote the
total number of photons detected by the IP-SPAD pixel during its fixed exposure
time, and let $X_i$ $(1\leq i \leq N_T)$ denote the timestamp of the
$i^\text{th}$ photon detection.  The measured inter-photon times
between successive photons is defined as $Y_i := X_{i+1} -
X_{i} - \taud$ (for $1 \leq i \leq N_T-1$). Note that $Y_i$'s follow an exponential
distribution. It is tempting to derive a closed-from maximum
likelihood photon flux estimator $\widehat\Phi$ for the true flux $\Phi$
using the log-likelihood function of the measured inter-photon times $Y_i$:
\begin{align}
\vspace{-0.15in}
  \log l(q\Phi; Y_1,\ldots,Y_{N_T-1}) = \log \left( \prod_{i=1}^{N_T-1}
                     q\Phi\,e^{-q\Phi Y_i} \right) \nonumber \\
  = - q\Phi \left( \sum_{n=1}^{N_T-1} Y_i \right) + (N_T-1)
                    \log q\Phi, \label{eq:loglik}
\vspace{-0.15in}
\end{align}
where $0\!\!<q<\!\!1$ is the quantum efficiency of the IP-SPAD pixel.
The maximum likelihood estimate $\widehat\Phi$ of the true photon flux is
computed by setting the derivative of Eq.~(\ref{eq:loglik}) to zero and
solving for $\Phi$:
\begin{equation}
\boxed{
  \widehat{\Phi} = \frac{1}{q}\frac{N_T-1}{X_{N_T}-X_1-(N_T-1)\taud}.
  \label{eq:flux_estimator}
}
\end{equation}

Although the above proof sketch captures the intuition of our flux
estimator, it leaves out two details. First, the total number of photons
$N_T$ is itself a random variable. Second, the times of capture of
future photons are constrained by the timestamps of preceding photon arrivals
because we operate in a finite exposure time $T$. The sequence of timestamps $Y_i$
cannot be treated as independent and identically distributed. \atul{In fact, it is
possible to show that the sequence of timestamps forms a Markov chain
\cite{rapp2019dead} where the conditional distribution of the $i^\text{th}$
inter-photon time conditioned on the previous inter-photon times is given by}:
\begin{equation}
p_{Y_i | Y_{1},\dots,Y_{i-1}}(t)=\begin{cases}
q\Phi e^{-q\Phi t} & 0< t< T_i\\
e^{-q\Phi T_i} \delta(t-T_i) & t = T_i\\
0 & \mbox{otherwise.}
\end{cases}\nonumber 
\end{equation}
Here $\delta(\cdot)$ is the Dirac delta function.  The $T_i$'s model the
shrinking effective exposure times for subsequent photon detections.
$T_1 = T$ and for $i>1$, 
$T_i = \max(0,T_{i-1} - Y_{i-1}-\taud)$.
The log-likelihood function can now be written as:
\begin{align*}
  \log l(q\Phi; Y_1,\ldots,Y_{L}) = \log \left( \prod_{i=1}^{\lceil\nicefrac{T}{\taud}\rceil} p_{Y_i | Y_{1},\dots,Y_{i-1}}(t) \right) \\
  \!\!= - q\Phi \max\left(\sum_{i=1}^{N_T}Y_i,T\!-\!N_T\taud\right) + N_T \log q\Phi.
\end{align*}
As shown in \nolink{\ref{suppl:MLE_conditional_derivation}} this likelihood function also
leads to the flux estimator given in Eq.~(\ref{eq:flux_estimator}).

We make the following key observations about the IP-SPAD flux estimator.
First, note that the estimator is only a function of the first and the last
photon timestamps, the exact times of capture of the intermediate photons do
not provide additional information.\footnote{As we show later in our hardware
implementation, in practice, it is useful to capture intermediate photon
timestamps as they allow us to calibrate for various pixel non-idealities.}
This is because photon arrivals follow Poisson statistics: the time until the
next photon arrival from the end of the previous dead-time is independent of
all preceding photon arrivals. Secondly, we note that the denominator in
Eq.~(\ref{eq:flux_estimator}) is simply the total time the IP-SPAD spends
waiting for the next photon to be captured while not in dead-time. Intuitively,
if the photon flux were extremely high, we will expect to see a photon
immediately after every dead-time duration ends, implying the denominator in
Eq.~(\ref{eq:flux_estimator}) approaches zero, hence $\widehat{\Phi}
\rightarrow \infty$.

\begin{figure}[!t]
	\centering \includegraphics[width=0.95\columnwidth]{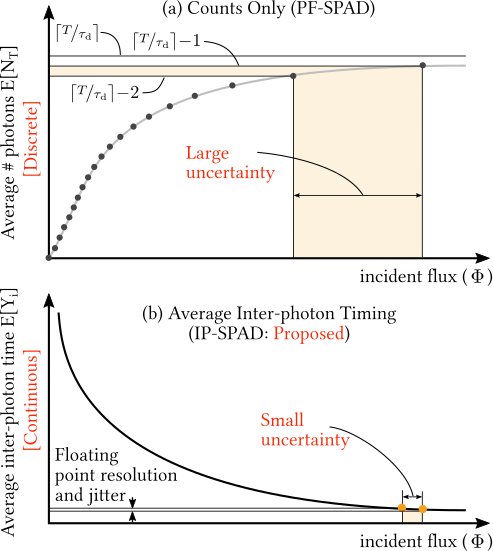}
	\caption{\textbf{Advantage of using photon timing over photon counts:}
	(a) Photon counts are inherently discrete. At high flux levels, even a
	small $\pm 1$ change in photon counts corresponds to a large flux uncertainty.
	(b) Inter-photon timing is inherently continuous. This leads to smaller
	uncertainty at high flux levels. The uncertainty depends on jitter and
	floating point resolution of the timing electronics. \label{fig:timing_quantization_noise}}
	\vspace{-0.15in}
\end{figure}

\begin{figure}[!t]
\centering \includegraphics[width=\columnwidth]{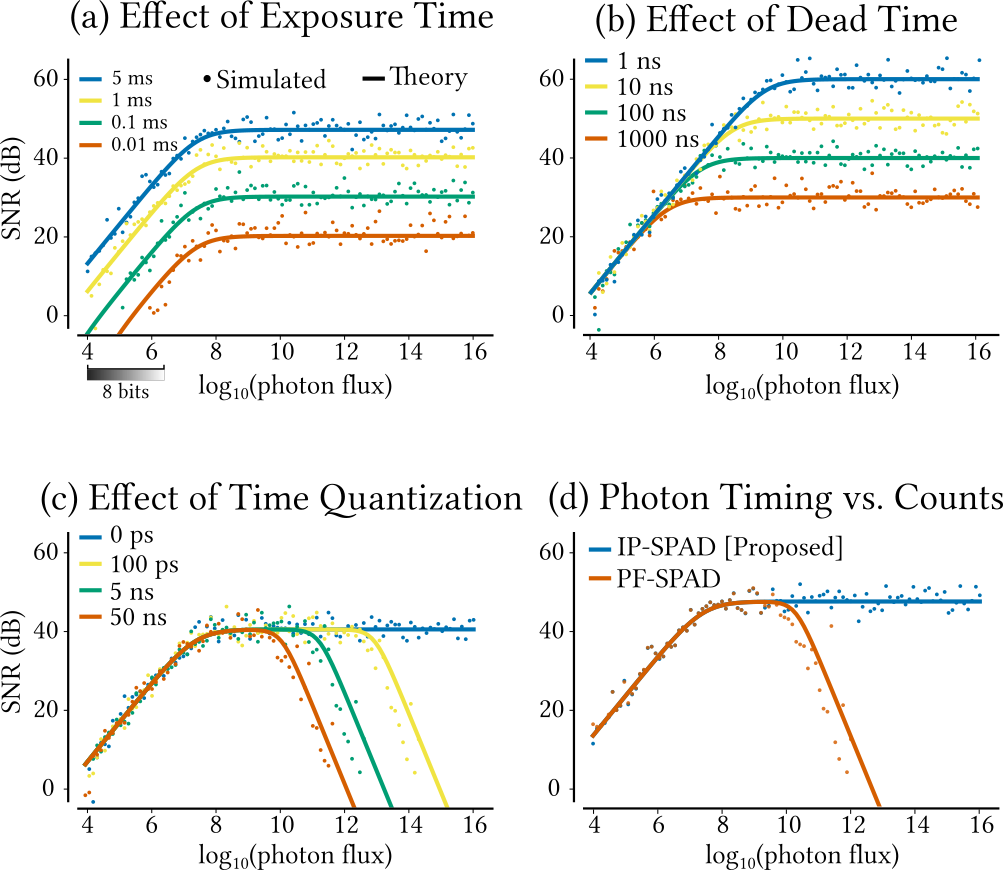}
\caption{\textbf{Effect of various IP-SPAD parameters on SNR:} We vary
different parameters to see the effect on SNR. The solid lines are theoretical
SNR curves while each dot represents a SNR average from 10 Monte Carlo
simulations. Unless otherwise noted the parameters used are $T=1$~\si{\ms},
$\taud=100$~\si{\ns}, $q=0.4$, and $\Delta=0$.  (a) As exposure time increases,
SNR increases at all brightness levels. (b) Decreasing the dead-time increases
the maximum achievable SNR, but provides little benefit in low flux.
(c) Coarser time quantization causes SNR drop-off at high flux values. (d) Our
IP-SPAD flux estimator outperforms a counts-only (PF-SPAD) flux estimator
\cite{ingle2019high} at high flux levels.\label{fig:snr}}
\vspace{-0.15in}
\end{figure}

\subsection{Sources of Noise}
Although, theoretically, the IP-SPAD scene brightness estimator in
Eq.~(\ref{eq:flux_estimator}) can recover the entire range of incident photon
flux levels, including very low and very high flux values, in practice, the
accuracy is limited by various sources of noise. To assess the performance of
this estimator, we use a signal-to-noise ratio (SNR) metric defined as
\cite{yang2011bits,ingle2019high}: 
\begin{equation}
  \mathsf{SNR}(\Phi) = 10 \log_{10} \left( \frac{\Phi^2}{\mathbf{E}[(\Phi-\widehat\Phi)^2]}\right) \label{eq:snr}
\end{equation}
Note that the denominator in Eq.~(\ref{eq:snr}) is the mean-squared-error
of the estimator $\widehat\Phi$, and is equal to the sum of the bias-squared
terms and variances of the different sources of noise.
The \emph{dynamic range} (DR) of the sensor is defined as the range of
brightness levels for which the SNR stays above a minimum specified threshold.
At extremely low flux levels, the dynamic range is limited due to IP-SPAD dark
counts which causes spurious photon counts even when no light is incident
on the pixel. This introduces a bias in $\widehat\Phi$. Since photon arrivals
are fundamentally stochastic (due to the quantum nature of light), the
estimator also suffers from Poisson noise which introduces a non-zero variance
term. Finally, at high enough flux levels, the time discretization $\Delta$
used for recording timestamps with the IP-SPAD pixel limits the maximum usable
photon flux at which the pixel can operate. Fig.~\ref{fig:sources_of_noise}(a)
shows the theoretical expression for bias and variance introduced by shot noise,
quantization noise and dark noise in an IP-SPAD pixel along with corresponding
expressions for a conventional image sensor pixel and a PF-SPAD pixel.
Fig.~\ref{fig:sources_of_noise}(b) shows example plots for these theoretical
expressions. For realistic values of $\Delta$ in the 100's of picoseconds range,
the IP-SPAD pixel has a smaller quantization noise term that allows reliable
brightness estimation at much higher flux levels than a PF-SPAD pixel. (See
\nolink{\ref{suppl:quantization}}).

\begin{figure}[!ht]
  \centering \includegraphics[width=0.90\linewidth]{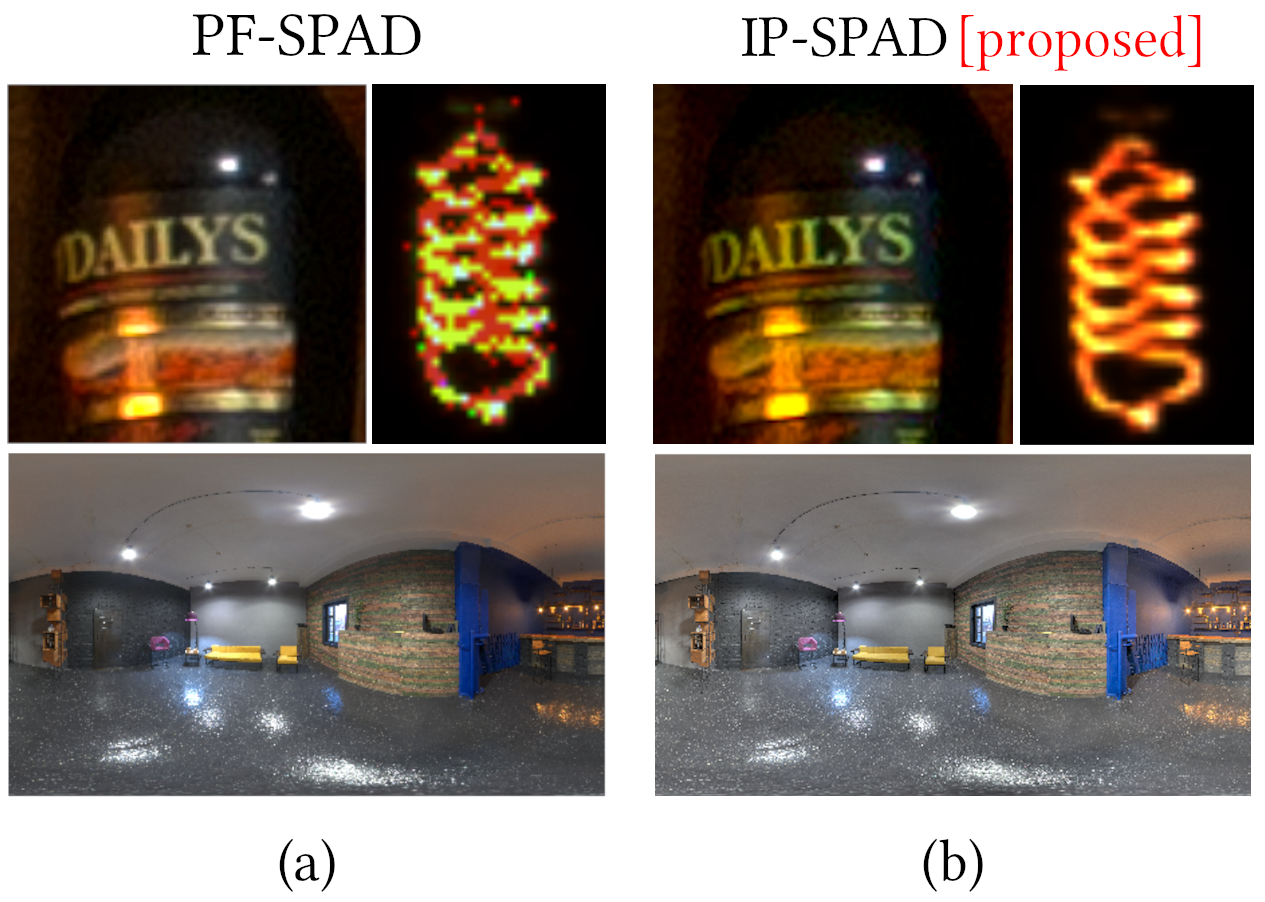}
  \caption{\textbf{Simulated HDR scene captured with a PF-SPAD (counts only)
  vs. IP-SPAD (inter-photon timing):}
  (a) Although a PF-SPAD can capture this extreme dynamic range scene in a
  single \SI{5}{\ms} exposure, extremely bright pixels such as the bulb
  filament that are beyond the soft-saturation limit appear noisy.
  (b) An IP-SPAD camera captures both dark and bright regions in a single
  exposure, including fine details around the bright bulb filament.
  In both cases, we set the SPAD pixel's quantum efficiency to 0.4, dead-time
  to \SI{150}{\ns} and an exposure time of \SI{5}{\ms}. The IP-SPAD has a time
  resolution of $\Delta=\SI{200}{\ps}$.  (Original image from
  HDRIHaven.com)\label{fig:sim_hdr}}
\end{figure}


\smallskip
\noindent{\bf Quantization Noise in PF-SPAD vs. IP-SPAD:}
Conventional pixels are affected by quantization in low flux and hard
saturation (full-well capacity) limit in high flux.  In contrast, a PF-SPAD
pixel that only uses photon counts is affected by quantization noise \emph{at
extremely high flux levels due to soft-saturation} \cite{ingle2019high}. This
behavior is unique to SPADs and is quite different from conventional sensors. A
counts-only PF-SPAD pixel can measure at most $\lceil
\nicefrac{T}{\tau_\text{d}}\rceil$ photons where $T$ is the exposure time and
$\tau_\text{d}$ is the dead-time \cite{ingle2019high}.  Due to a non-linear
response curve, as shown in Fig.~\ref{fig:timing_quantization_noise}(a), a
small change of $\pm 1$ count maps to a large range of flux values. Due to the
inherently discrete nature of photon counts, even a small fluctuation (due to
shot noise or jitter) can cause a large uncertainty in the estimated flux.


The proposed IP-SPAD flux estimator uses timing information which is inherently
continuous.  Even at extremely high flux levels, photon arrivals are random and
due to small random fluctuations, the time interval between the first and last
photon ($X_{N_T}-X_1$) is not exactly equal to $T$.
Fig.~\ref{fig:timing_quantization_noise}(b) shows the intuition for why
fine-grained inter-photon measurements at high flux levels can enable flux
estimation with a smaller uncertainty than counts alone. In practice, the
improvement in dynamic range compared to a PF-SPAD depends on the time
resolution, which is limited by hardware constraints like floating point
precision of the TDC electronics and timing jitter of the SPAD pixel.
Simulations in Fig.~\ref{fig:snr} suggest that even with a \SI{100}{\ps} time
resolution the 20-dB dynamic range improves by 2 orders of magnitude over using
counts alone.

\smallskip
\noindent{\bf Single-Pixel Simulations:}
We verify our theoretical SNR expression using single-pixel Monte Carlo
simulations of a single IP-SPAD pixel. For a fixed set of parameters we run 10
simulations of an IP-SPAD at 100 different flux levels ranging $10^4-10^{16}$
photons per second. Fig. \ref{fig:snr} shows the effect of various pixel
parameters on the SNR. The overall SNR can be increased by either
increasing the exposure time $T$ or decreasing the dead-time $\taud$;
both enable the IP-SPAD pixel to capture more total
photons. The maximum achievable SNR is theoretically equal to
$10\log_{10}\left(\nicefrac{T}{\taud}\right)$. The IP-SPAD
SNR degrades at high flux levels due because photon timestamps cannot be
captured with infinite resolution.  A larger floating point quantization bin
size $\Delta$ increases the uncertainty in photon timestamps. If the time bin
is large enough, there is no advantage in using the timestamp-based brightness
estimator and the performance reverts to a counts-based flux estimator
\cite{Antolovic_2018, ingle2019high}.

\section{Results}
\subsection{Simulation Results}
\noindent{\bf Simulated RGB Image Comparisons:}
Fig.~\ref{fig:sim_hdr} shows a simulated HDR scene with extreme brightness
variations between the dark text and bright bulb filament. We use a \SI{5}{\ms}
exposure time for this simulation. The PF-SPAD and
IP-SPAD both use pixels with $q=0.4$ and $\taud=\SI{150}{\ns}$.
The PF-SPAD only captures photon
counts whereas the IP-SPAD captures counts and timestamps with a resolution
of $\Delta=\SI{200}{\ps}$. Notice that the extremely bright pixels on the bulb
filament appear noisy in the PF-SPAD image. This degradation in SNR at high
flux levels is due to its soft-saturation phenomenon. The IP-SPAD, on the other
hand, captures the dark text and noise-free details in the bright bulb filament
in a single exposure. Please see supplement for additional comparisons with a
conventional camera.


\begin{figure*}[!ht]
  \centering \includegraphics[width=0.99\linewidth]{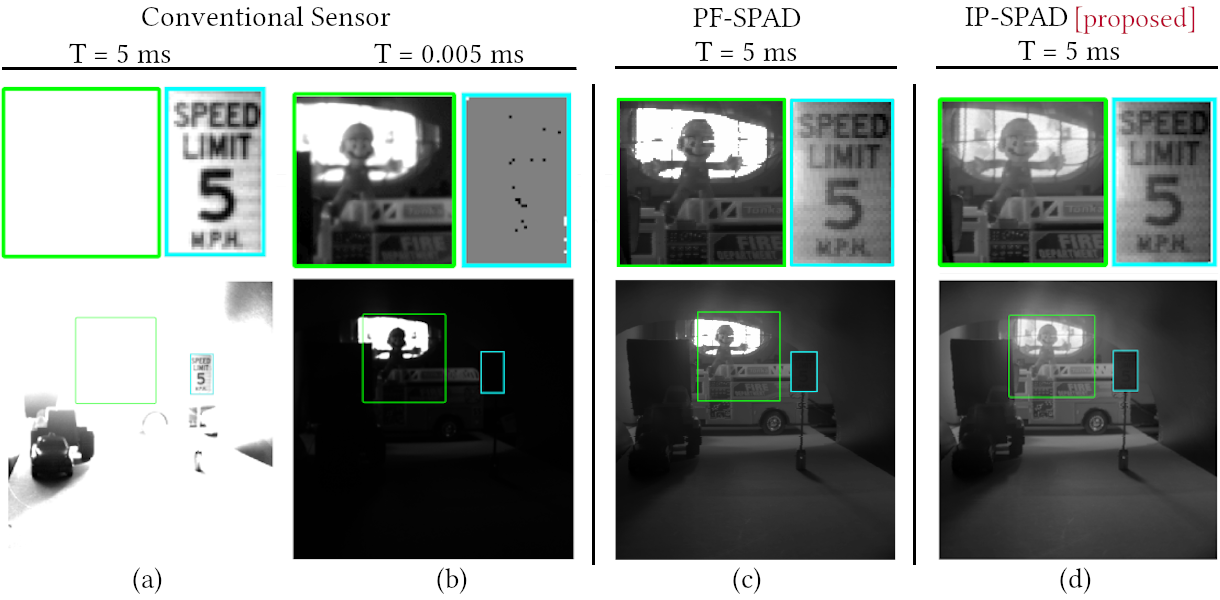}
  \caption{\textbf{Experimental ``Tunnel'' scene:}  (a-b) Images from a
  conventional sensor with long and short exposure times. Notice that both the speed
  limit sign and the toy figure cannot be captured simultaneously with a single
	exposure. Objects outside the tunnel appear saturated even with the
	shortest exposure time possible with our CMOS camera. (c) A PF-SPAD
  \cite{ingle2019high} only uses photon counts when estimating flux.
  Although it captures much higher dynamic range than the
  conventional CMOS camera, the bright pixels near the halogen lamp 
  appear saturated. (d) Our IP-SPAD single-pixel hardware prototype
  captures both the dark and the extremely bright regions with a single
  exposure. Observe that the fine details within the halogen
  lamp are visible. \label{fig:tunnel}}
  \vspace{-0.15in}
\end{figure*}

\subsection{Single-Pixel IP-SPAD Hardware Prototype}
\begin{figure}[!ht]
  \centering
  \includegraphics[width=\columnwidth]{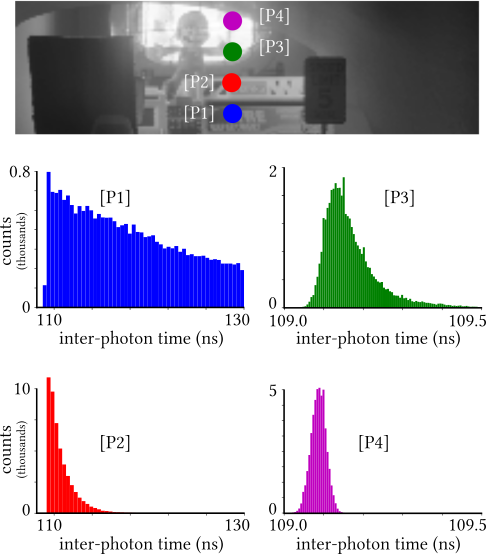}
\caption{\textbf{Rise-time Non-ideality in Measured IP-SPAD Histograms:} We show
four inter-photon histograms for pixels in the experimental ``Tunnel'' scene.
The histograms of [P1] and [P2] have an ideal exponentially decaying shape.
However, at the brighter points, [P3] and [P4], the inter-photon histograms
deviate from an ideal exponential shape. This is because the IP-SPAD pixels requires
$\sim \SI{100}{\ps}$ rise time to re-activate after the end of the previous dead-time.
  \vspace{-0.25in}
\label{fig:exp_hists}}
\end{figure}

Our single-pixel IP-SPAD prototype is a modified version of a fast-gated SPAD
module \cite{Buttafava2014}. Conventional dead-time control circuits for a SPAD
rely on digital timers that have a coarse time-quantization limited by the
digital clock frequency and cannot be used for implementing an IP-SPAD. We
circumvent this limitation by using coaxial cables and low-jitter voltage
comparators to generate ``analog delays'' that enable precise control of the
dead-time with jitter limited to within a few \si{\ps}. We used a \SI{20}{\m}
long co-axial cable to get a dead-time of \SI{110}{\ns}. The measured dead-time
jitter was $\sim\SI{200}{\ps}$. This is an improvement over previous PF-SPAD
implementations \cite{ingle2019high} that relied on a digital timer circuit
whose time resolution was limited to $\sim\SI{6}{\ns}$.

The IP-SPAD pixel is mounted on two translation stages that raster scan the
image plane of a variable focal length lens. The exposure time per pixel
position depends on the translation speed along each scan-line.  We capture
400$\times$400 images with an exposure time of \num{5}~\si{\ms} per pixel
position. The total capture takes $\sim\!\!\!15$ minutes. Photon timestamps are
captured with a \SI{1}{\ps} time binning using a time-correlated single-photon
counting (TCSPC) system (PicoQuant HydraHarp400). A monochrome camera
(PointGrey Technologies GS3-U3-23S6M-C) placed next to the SPAD captures
conventional camera images for comparison.  The setup is arranged carefully to
obtain approximately the same field of view and photon flux per pixel for both
the IP-SPAD and CMOS camera pixels.


\begin{figure}[!htb]
  \centering
  \includegraphics[width=0.9\columnwidth]{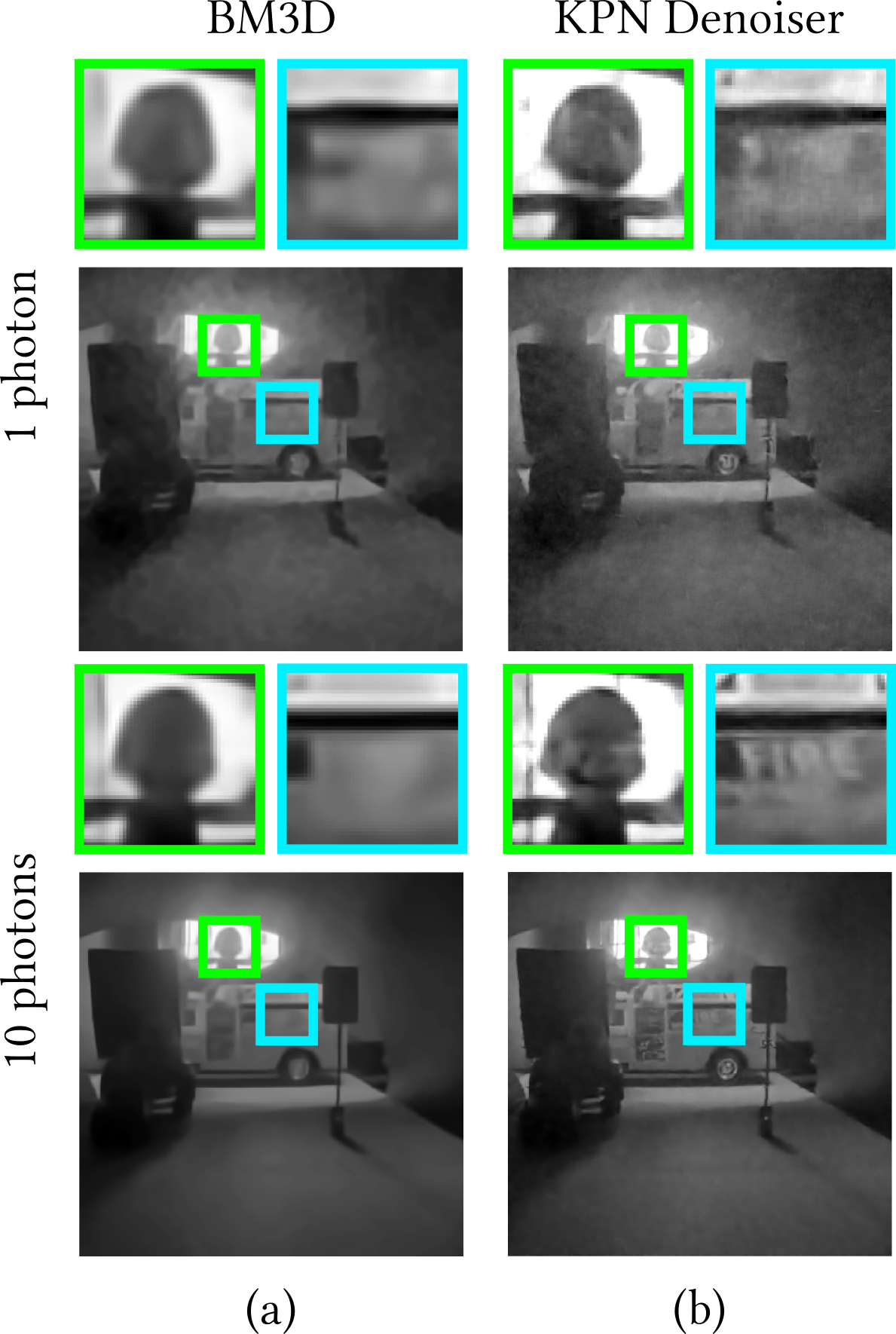}
	\caption{\textbf{IP-SPAD Imaging in Low Photon Count Regime:}
  This figure shows IP-SPAD images captured with very few photons and denoised
  with two different methods: (a) an off-the-shelf BM3D denoiser, and (b) a DNN
  denoiser based on a kernel prediction network architecture.
  Details like the text on the fire-truck are visible with as few as 10 photons per
  pixel.\label{fig:low_counts}}
  \vspace{-0.15in}
\end{figure}

\subsection{Hardware Experiment Results}
\noindent\textbf{HDR Imaging:}
Fig.~\ref{fig:tunnel} shows an experiment result using our single-pixel
raster-scanning hardware prototype. The ``Tunnel'' scene contains dark objects
like a speed limit sign inside the tunnel and an extremely bright region
outside the tunnel from a halogen lamp.  This scene has a dynamic range of over
\num[retain-unity-mantissa=false]{1e7}:\num{1}. The conventional CMOS camera
(Fig.~\ref{fig:tunnel}(a-b)), requires multiple exposures to cover this dynamic
range.  Even with the shortest possible exposure time of \SI{0.005}{\ms}, the
halogen lamp appears saturated. Our IP-SPAD prototype captures details of both
the dark regions (text on the speed limit sign) simultaneously with the bright
pixels (outline of halogen lamp tube) in a single exposure.
Fig.~\ref{fig:tunnel}(c) and (d) shows experimental comparison between a
PF-SPAD (counts-only) image \cite{ingle2019high} and the proposed IP-SPAD image
that uses photon timestamps. Observe that in extremely high flux levels (in
pixels around the halogen lamp) the PF-SPAD flux estimator fails due to the
inherent quantization limitation of photon counts. The IP-SPAD preserves 
details in these extremely bright regions, like the shape of the halogen 
lamp tube and the metal cage on the lamp.

\smallskip
\noindent\textbf{Hardware Limitations:}
The IP-SPAD pixel does not exit the dead-time duration instantaneously and in
practice it takes around \SI{100}{\ps} to transition into a fully-on state.
Representative histograms for four different locations in the experiment tunnel
scene are shown in Fig.~\ref{fig:exp_hists}. Observe that at lower flux levels
(pixels [P1] and [P2]) the inter-photon histograms follow an exponential
distribution as predicted by the Poisson model for photon arrival statistics.
However, at pixels with extremely high brightness levels (pixels [P3] and [P4]
on the halogen lamp), the histograms have a rising edge denoting the transition
phase when the pixel turns on after the end of the previous dead-time.
We also found that in practice the dead-time is not constant and exhibits a
drift over time (especially at high flux values) due to internal heating.  Such
non-idealities, if not accounted for, can cause uncertainty in photon timestamp
measurements, and limit the usability of our flux estimator in the high photon
flux regime.  Since we capture timestamps for every photon detected in a fixed
exposure time, it is possible to correct these non-idealities in
post-processing by estimating the true dead-time and rise-time from these
inter-photon timing histograms. See \nolink{\ref{suppl:dead time drift}} for
details.

\smallskip
\noindent\textbf{IP-SPAD Imaging with Low Photon Counts:}
The results so far show that precise photon timestamps from an IP-SPAD
pixel enables imaging at extremely high photon flux levels. We now show that it
is also possible to leverage timing information when the IP-SPAD pixel captures
very few photons per pixel. We simulate the low photon count regime by keeping
the first few photons and discarding the remaining photon timestamps for each
pixel in the experimental ``Tunnel'' scene. Fig.~\ref{fig:low_counts} shows
IP-SPAD images captured with as few as 1 and 10 photons per pixel and denoised
using an off-the-shelf BM3D denoiser and a deep neural network denoiser that
uses a kernel prediction network (KPN) architecture \cite{burstkpn_2018}. We
can recover intensity images with just one photon timestamp per pixel using
real data captured by our IP-SPAD hardware prototype.  Quite remarkably, with
as few as 10 photons per pixel, image details such as facial features and text
on the fire truck are discernible. Please see
\nolink{\ref{suppl:timing_usefulness}} for details about the KPN denoiser and
\nolink{\ref{suppl:additional_results}} for additional experimental results and
comparisons with other denoising algorithms.

\section{Future Outlook}
The analysis and experimental proof-of-concept shown in this paper were
restricted to a single IP-SPAD pixel. Recent advances in CMOS SPAD technology
that rely on 3D stacking \cite{Henderson_2019_ISSCC} can enable larger arrays
of SPAD pixels for passive imaging. This will introduce additional design
challenges and noise sources not discussed here. In
\nolink{\ref{suppl:pixel_designs}} we show some pixel architectures for an
IP-SPAD array that could be implemented in the future.

Arrays of single-photon image sensor pixels are being increasingly used for 2D
intensity imaging and 3D depth sensing
\cite{yoshida_2020,Lee_2020,Mardirosian_2020} in commercial and consumer
devices. When combined with recent advances in high-time-resolution SPAD sensor
hardware, the methods developed in this paper can enable extreme imaging
applications across various applications including consumer photography, vision
sensors for autonomous driving and robotics, and biomedical optical imaging.




{\small
\bibliographystyle{ieee_fullname}
\bibliography{sample-base}
}

%
%

\clearpage
\onecolumn
\normalsize
\renewcommand{\figurename}{Supplementary Figure}
\captionsetup[figure]{name={Suppl. Fig.}}
\renewcommand{\thesection}{Supplementary Note \arabic{section}}
\renewcommand{\theequation}{S\arabic{equation}}
\setcounter{figure}{0}
\setcounter{section}{0}
\setcounter{equation}{0}
\setcounter{page}{1}
\renewcommand*{\thefootnote}{$\dagger$}

\begin{center}
\huge Supplementary Document for\\
\huge ``\mytitle'' \\[0.7cm]
\large Atul Ingle$^\ast$, Trevor Seets, Mauro Buttafava, Shantanu Gupta, \\
Alberto Tosi, Mohit Gupta$^\dagger$, Andreas Velten\footnote{Equal contribution.\\
$^\ast$Email: ingle@uwalumni.com}\\[0.5cm]
(CVPR 2021)
\end{center}

\renewcommand*{\thefootnote}{\arabic{footnote}}

\section{Image Formation}\label{suppl:MLE_conditional_derivation}

\begin{figure}[!ht]
\centering \includegraphics[width=0.65\textwidth]{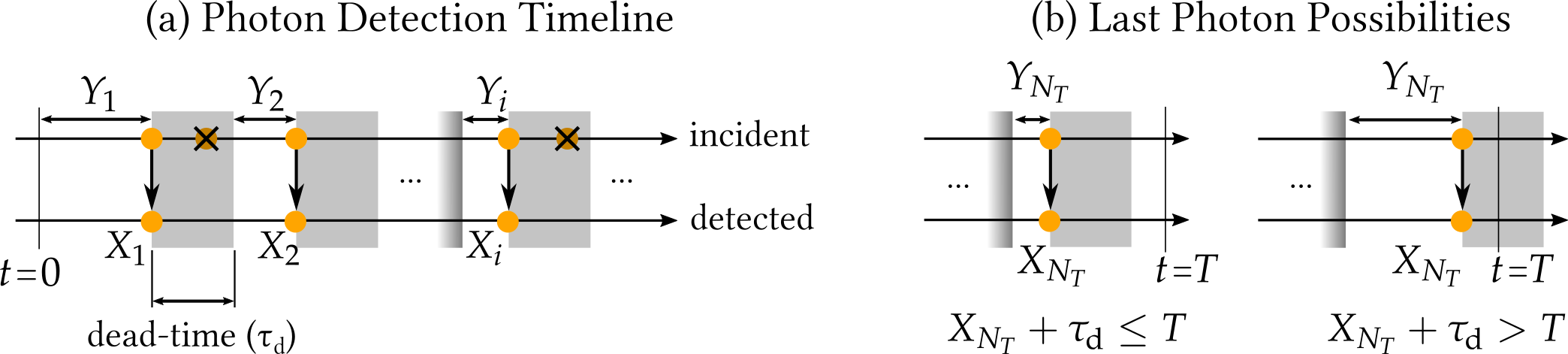}
\caption{\textbf{Photon Detection Timeline:} (a) The photon timeline shows the
random variables used in the derivation of our photon flux estimator.  $X_i$'s
denote the photon arrival time with respect to the start of the exposure at
$t=0$ and $Y_i$'s denote the $i^{th}$ time-of-darkness. (b) There are two
possibilities for the final dead-time window after the last photon detection.
In high photon flux scenarios, the final dead-time ends after the end of the
exposure time $(X_{N_T}+\taud > T)$ with high probability.
\label{suppl:timeline}}
\end{figure}

Consider an IP-SPAD sensor pixel with quantum efficiency $q$ exposed to a
photon flux of $\Phi$ photons/second. Photon arrivals follow a Poisson process,
so inter-photon times follow an exponential distribution with rate $q\Phi$.
After a detection event the IP-SPAD is unable to detect photons for a period of
$\taud$. Because of the memoryless property of a Poisson process, the arrival
time of a photon after the end of a dead-time window (called the
\textit{time-of-darkness}), follows an exponential distribution with rate
$q\Phi$. The probability that no photons are detected in the exposure time $T$
is equal to $\int^\infty_T q\Phi e^{-q\Phi t} dt =e^{-q\Phi T}.$

Let the first time-of-darkness be denoted by $Y_1$. If no photons are detected,
we define $Y_1:=T$. If $Y_1<T$ it follows an exponential distribution.
Therefore, the probability density function of $Y_1$ can be written as:

\begin{equation}
Y_{1}{\sim}f_{Y_{1}}(t)=\begin{cases}
q\Phi e^{-q\Phi t} & \mbox{for } 0< t< T\\
e^{-q\Phi T} \delta(t-T)&\mbox{for } t=T\\
0 & \mbox{otherwise,}
\end{cases} \label{eq:first_photon_dist}
\end{equation}
where $\delta$ is the Dirac delta function.

Now consider $Y_2$, the second time-of-darkness. $Y_2$ is non-zero if and only
if $Y_1 \neq T$ (to be the second, there must be a first). If the second photon
is detected, $Y_2$ will be exponentially distributed. But the exposure time
interval shrinks because a time interval of $Y_1+\taud$ has elapsed due to
the first photon detection.  We define the remaining exposure time
$T_2=\max(0,T-Y_1-\tau_d$), where the $\max()$ function ensures $T_2$ is
non-negative. Then the probability distribution of $Y_2$ conditioned on $Y_1$
will be given by replacing $T$ for $T_2$ in Eq.~(\ref{eq:first_photon_dist}).
More generally, the conditional distribution of $Y_i$ can be written as:

\begin{equation}
Y_{i} \sim f_{Y_i | Y_{1}\dots Y_{i-1}}(t | Y_{1}\dots Y_{i-1})=\begin{cases}
q\Phi e^{-q\Phi t} & \mbox{for } 0< t < T_i\\
e^{-q\Phi T_i} \delta(t-T_i) &\mbox{for } t=T_i\\
0 & \mbox{otherwise,}
\end{cases} \label{eq:interarrival_distr}
\end{equation}
where,
\begin{align}
T_1 &=T\nonumber\\
T_i &= \max(0,T_{i-1} - Y_{i-1}-\taud))\nonumber\\
 &=\max\left(0,T -\sum^{i-1}_{j=1}(Y_j+\taud)\right).
\end{align}

The $T_i$'s model the fact that the effective exposure time for the $i^{th}$
photon shrinks due to preceding photon detections. Note if $Y_i = T_i$ then no
$i^\text{th}$ photon is detected and $Y_{i+1}=T_{i+1}=0$. Note that the $X_i$
in the main text are related to $Y_i$ by $X_i-X_{i-1}-\taud=:Y_i$ for $i\geq2$
and $X_1=Y_1$. Suppl.  Fig.~\ref{suppl:timeline}(a) shows  $X_i$ and $Y_i$ on a
photon timeline.

\subsection*{Maximum Likelihood Flux Estimator}
For a fixed exposure time $T$, the maximum number of possible photon detections
is $L=\big\lceil\frac{T}{\taud}\big\rceil$. Let $N$ be the number of detected
photons, then $Y_{N+1}$ will be the last possibly non-zero time-of-darkness,
and $Y_{N+2}\dots Y_L =0$ with probability $1$. The log-likelihood of the
unknown flux value given the set of time-of-darkness measurements $Y_1\dots
Y_L$ is given by:

\begin{align}
  \log l(q\Phi; Y_1,\ldots,Y_{L}) &= \log \left( \prod_{i=1}^{L} f_{Y_{i}}(Y_i | Y_{1}\dots Y_{i-1}) \right) \nonumber \\
  &= \log \left(f_{Y_{N+1}}(Y_{N+1} | Y_{1}\dots Y_{N}) \prod_{i=1}^{N} f_{Y_{i}}(Y_i | Y_{1}\dots Y_{i-1}) \right) \nonumber \\
  &= \log \left( e^{-q\Phi T_{N+1}}\prod_{i=1}^{N} q\Phi\,e^{-q\Phi Y_i} \right) \nonumber \\
  &= - q\Phi \left( T_{N+1}+\sum_{i=1}^{N}Y_i \right) + N\log q\Phi \nonumber \\
  &= - q\Phi\,\left( \max\left(0,T-\sum_{i=1}^{N}Y_i-\taud\right)+\sum_{i=1}^{N}Y_i\right) + N \log q\Phi\nonumber\\
  &= - q\Phi\, \max\left(\sum_{i=1}^{N}Y_i,T-N\taud\right) + N \log q\Phi.
  \label{eq:loglik_cond}
\end{align}
We find the maximum likelihood estimate, $\widehat{\Phi}$, by setting the
derivative of Eq.(\ref{eq:loglik_cond}) to zero and solving for $\Phi$:

\begin{align}
  - q\max\left(\sum_{i=1}^{N}Y_i,T-N\taud\right) + \frac{N}{\widehat\Phi} &= 0,
\end{align}
which gives: 
\begin{align}
  \widehat\Phi=\frac{N}{q\max\left(\sum_{n=1}^{N}Y_i,T-N\taud\right)}.\label{eq:MLE_cond}
\end{align}

The $\max()$ function can be thought of as selecting the time-of-darkness based
on whether or not the final dead-time window ends after $t=T$, see Suppl.
Fig.~\ref{suppl:timeline}(b). In practice the beginning and end of the exposure
time may not be known precisely, introducing uncertainty in $X_1$ and $T$.
Because of this we instead use an approximation:
\begin{align}
  \widehat\Phi=\frac{N-1}{q\sum_{n=2}^{N}Y_i}.\label{eq:MLE_cond_simp}
\end{align}
Plugging in $Y_i = X_i - X_{i-1} - \taud$ gives Eq.~(\ref{eq:flux_estimator})
in the main text.


\subsection*{Flux Estimator Variance}
Let $N$ be the number of photons detected in an exposure time $T$. Using the
law of large numbers for renewal processes we find the expectation and the
variance of $N$ to be:

\begin{align}
  \text{E}[N] = \frac{q\Phi(T+\taud)}{1+q\Phi \taud} \\
  \text{Var}[N]=\frac{q\Phi(T+\taud)}{(1+q\Phi \taud)^3}
\end{align}

In the following derivation we will assume $N$ is large enough that it can be
assumed to be constant for a given $T$. This holds, for example, when $\Phi \gg
\frac{1}{T}$.  This assumption also allows us to approximate $Y_i$'s as i.i.d.
shifted exponential random variables. We will consider the estimator in
Eq.~(\ref{eq:MLE_cond_simp}) where the sum in the denominator is given by
$S_{N_T}=Y_2+Y_3+...Y_N$ and letting $N_T=N-1$.  The final photon timestamp
$S_{N_T}$ is the sum of exponential random variables and follows a gamma
distribution:

\begin{equation}
S_{N_T}\stackrel{ }{\sim}f_{S_{N_T}}(t)=\begin{cases}
\frac{(q\Phi)^{N_T} t^{N_T-1} e^{-q\Phi t}}{(N_T-1)!} & \mbox{for } t\geq0\\
0 & \mbox{otherwise.}
\end{cases} \label{eq:gamma_interarrival_distr}
\end{equation}

The mean of $\widehat\Phi$ can be computed as: 
\begin{align}
  \text{E}\left[\frac{N_T}{q S_{N_T}}\right]&= \frac{N_T}{q}\text{E}\left[\frac{1}{S_{N_T}}\right] \nonumber\\
  &=  \frac{N_T}{q}\int^{\infty}_{0} \frac{(q\Phi)^{N_T} t^{N_T-1} e^{-q\Phi t}}{t(N_T-1)!}\, dt \nonumber\\
  &=  \frac{N_T}{q}\frac{q\phi}{N_T-1}\int^{\infty}_{0} \frac{(q\Phi)^{N_T-1} t^{N_T-2} e^{-q\Phi t}}{(N_T-2)!}\, dt\nonumber\\
  &= \frac{N_T}{N_T-1}\Phi \label{eq:bias}
\end{align}
where the last line comes from the recognizing the argument of the integral as
the p.d.f. for a gamma distribution and for large $N_T$, $\frac{N_T}{N_T-1}
\approx 1$.

The second moment of $\widehat\Phi$ is given by:
\begin{align}
  \text{E}\left[\left(\frac{N_T}{q S_{N_T}}\right)^2\right] 
  &=\frac{N_T^2}{q^2}\text{E}\left[\frac{1}{S_{N_T}^2}\right] \nonumber\\
  &=\frac{N_T^2}{q^2}\int^{\infty}_{0} \frac{(q\Phi)^{N_T} t^{N_T-1} e^{-q\Phi t}}{t^2(N_T-1)!} dt\nonumber\\
  &=\frac{(q\Phi)^2N_T^2}{q^2(N_T-1)(N_T-2)}\int^{\infty}_{0} \frac{(q\Phi)^{N_T-2} t^{N_T-3} e^{-q\Phi t}}{(N_T-3)!} dt\nonumber\\
  &=\frac{\Phi^2 N_T^2}{(N_T-1)(N_T-2)}
\end{align}
This expression is valid for $N_T>2$. The variance of $\widehat\Phi$ is given by: 
\begin{align}
  \text{Var}\left[\frac{N_T}{q S_{N_T}}\right] &=\Phi^2\frac{N_T^2}{(N_T-2)(N_T-1)}-\frac{N_T^2}{(N_T-1)^2}\Phi^2 \nonumber \\
  &=\Phi^2\frac{N_T^2}{(N_T-2)(N_T-1)^2}\nonumber\\
  &\approx \Phi^2\frac{1}{N_T}\\
  &= \Phi^2 \frac{ q\Phi\taud+1}{q\Phi(T+\taud)}\\
  &\approx \Phi \frac{ q\Phi\taud+1}{q T}
\end{align}
where we replace $N_T$ with its mean value. The last line follows if we assume
$T\gg\taud$.
Finally, the SNR is given by:
\begin{align}
  \text{SNR} &= 20\log_{10}\frac{\Phi}{\sqrt{\text{Var}[\frac{N_T}{q S_{N_T}}]}} \nonumber\\
  &= 10\log_{10}\frac{q\Phi T}{ q\Phi \taud+1}
  \label{eq:SNRTiming}
\end{align}

We make the following observations about our estimator $\widehat\Phi$:
\begin{myitemize}
  \item At high flux, when $N_T$ is large enough, Eq.~(\ref{eq:bias}) reduces to
    $\text{E}[\widehat\Phi] = \Phi$, i.e.  our estimator is unbiased.
  \item Unlike \cite{ingle2019high} which only uses photon counts $N_T$, our
    derivation explicitly accounts for individual inter-photon timing
    information captured in $S_{N_T}$.
  \item As $\Phi\rightarrow\infty$,
    $\text{SNR}\rightarrow10\log_{10}(\frac{T}{\taud})$. So at high flux the
    SNR will flatten out to a constant independent of the true flux $\Phi$.  In
    practice, the SNR drops at high flux due to time quantization, discussed
    next in \ref{suppl:quantization}.
\end{myitemize}

\clearpage
\section{Time Quantization \label{suppl:quantization}}
Consider an IP-SPAD with quantum efficiency $q$, dead time $\taud$, and time
quantization $\Delta$ that detects photons over exposure time $T$. To match
our hardware prototype, the start of the dead time window is not quantized and
time stamps are quantized by $\Delta$. The quantization noise variance term
derived in previous work \cite{ingle2019high,Antolovic_2018} that relies on a
counts-only measurement model is given by:

\begin{equation}
V_\text{count-quantization} = \frac{(1+q\Phi \taud)^4}{12 q^2 T^2}.\label{suppl:count_quant} 
\end{equation}

We derive a modified quantization noise variance expression by modifying this
counts-only expression to account for two key insights gained from extensive
simulations of SNR plots for our timing-based IP-SPAD flux estimator. First, we
note that the timing-based IP-SPAD flux estimator follows a similar SNR curve
as the counts-based PF-SPAD flux estimator when  $\Delta=\taud$. Second, the
rate at which the SNR drop off moves slows after $\Delta$ exceeds $\taud$. In
this way we propose a new time quantization term:
\begin{equation}
V_\text{time-quantization} = \frac{(1+q\Phi \taud +q\Phi \Delta)^2(1+q\phi \Delta)^2}{12 q^2 T^2}.\label{suppl:quantization_noise} 
\end{equation}

Note we break the quartic term from Eq.~(\ref{suppl:count_quant}) into two
quadratic terms. The two quadratic terms strike a balance between quantization due
to counts and timing. If $\Delta=0$ then $V_\text{time quantization}$ is an order
2 polynomial with respect to $\Phi$ which leads to a constant SNR at high flux.
Also note if $\Delta=\taud$ the time quantization term is roughly equal to the
counts quantization term. We found this expression matches simulated IP-SPAD
SNR curves for a range of dead-times and exposure times.

\clearpage
\section{IP-SPAD Imaging with Low Photon Counts\label{suppl:timing_usefulness}}
The scene brightness estimator (Eq.~(\ref{eq:flux_estimator})) 
requires the IP-SPAD pixel to capture at least two photons; It
does not make sense to talk about ``inter-photon'' times with only one photon.
The situation where an IP-SPAD pixel captures only one incident photon
timestamp can be thought of as an extreme limiting case of passive inter-photon
imaging under low illumination. 

Intuitively, we can reconstruct an image from a single photon timestamp per
pixel by simply computing the reciprocal of the first photon timestamp at each
pixel. Brighter scene points should have a smaller first-photon timestamp (on
average) because, with high probability, a photon will be detected almost
immediately after the pixel starts collecting light.  In this supplementary
note we show that the conditional distribution of this first photon timestamp
(conditioned on there being at least one photon detection) is a uniform random
variable:
\[
  \{Y_1 | Y_1 \leq T\} \sim \mathcal{U}[0,T].
\]
when operating under low incident photon flux. This implies that timestamps
provide no additional information beyond merely the fact that at least one
photon was detected. We must, therefore, relax the requirement of a constant
exposure time and allow each pixel to capture at least one photon by allowing
variable exposure times per pixel. When operated this way, first-photon
timestamps do carry useful information about the scene brightness.  The
estimate of the scene pixel brightness is given by $\widehat \Phi =
\nicefrac{1}{q\,Y_i}$.

When the total number of photons is extremely small, the information contained
in the timestamp data is extremely noisy.  We leverage spatial-priors-based
image denoising techniques that have been developed for conventional images,
and adapt them denoising these noisy IP-SPAD images.  Coupled with the inherent
sensitivity of SPADs, this enables us to reconstruct intensity images with just
a single photon per pixel \cite{Johnson_2020}.

In this section we show that for passive imaging in the low photon flux regime
with a constant exposure time per pixel, the timestamp of the first arriving
photon is a uniform random variable and hence, carries no useful information
about the true photon flux. If we drop the constant exposure time constraint
and instead operate in a regime where each pixel is allowed to wait until the
first photon is captured (random exposure time per pixel), then the
first-photon timestamps carry useful information about the flux, albeit noisy.

\subsection{When Do Timestamps Carry Useful Information?}
\begin{figure}[!ht]
\centering \includegraphics[width=0.45\textwidth]{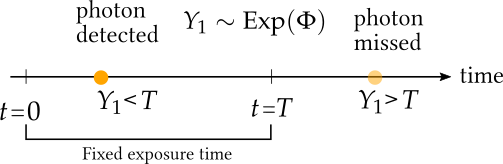}
\caption{We capture the first arriving photon and record
its arrival time in a fixed exposure time T. Note that in the
low photon flux regime $\Phi T \ll 1$, so there is a high
probability that zero photons are detected in the time interval $[0,T]$.
\label{fig:first_photon_timeline}}
\end{figure}

Let us assume an IP-SPAD pixel operating with a fixed exposure time $T$ is
observing a scene point with photon flux $\Phi$. We assume that the photon
flux is low enough so that the pixel captures at most one
photon during this exposure time. The (random) first-photon arrival time is
denoted by $Y_1$ as shown in Suppl. Fig.~\ref{fig:first_photon_timeline}. We
would like to know if the first photon time-of-arrival carries useful
information about $\Phi$.

We derive the probability distribution of $Y_1$, conditioning on $Y_1\leq T$.
For any $t>0$,
\begin{align}
     P(Y_1\leq t | Y_1\leq T) &= \frac{P(Y_1\leq t \cap Y_1 \leq T)}{P(Y_1\leq T)} \label{eq1}\\
       &= \frac{P(Y_1 \leq t)}{P(Y_1\leq T)}  \label{eq2}\\
       &= \frac{1-e^{-\Phi t}}{1-e^{-\Phi T}} \label{eq3}
\end{align}
where Eq.~(\ref{eq1}) follows from Bayes's rule,
Eq.~(\ref{eq2}) assumes $t\leq T$ (otherwise the answer is 1, 
trivially) and Eq.~(\ref{eq3}) is obtained by plugging
in the c.d.f. of $Y_1 \sim \text{Exp}(\Phi).$

Due to the low flux assumption, $\Phi \ll \frac{1}{T}$. Then $\Phi t \leq \Phi
T \ll 1$ and we can approximate $1-e^{-\Phi T} \approx \Phi T$ and $1-e^{-\Phi
t} \approx \Phi t$. This gives
\begin{equation}
    P(Y_1\leq t | Y_1\leq T) = \frac{t}{T}
\end{equation}
which is the c.d.f. of a uniform random variable. This implies that, in the low
photon flux regime the arrival time distribution converges weakly to a uniform
random variable:
\[
  \{Y_1 | Y_1 \leq T\} \overset{D}{\rightarrow} \mathcal{U}[0,T].
\]

For low illumination conditions, we drop the requirement of a fixed exposure
time and allow the IP-SPAD pixel to wait until the first photon timestamp
is captured.

\subsection{KPN-based Denoising Network for Low Light IP-SPAD Imaging}
In principle, any standard neighborhood-based image denoising algorithm (e.g.,
bilateral filtering \cite{paris2007gentle} and BM3D \cite{dabov2007image}) can
be applied to the IP-SPAD images captured in a low photon count regime. But the
heavy-tailed nature of the timestamps poses problems to off-the-shelf denoising
algorithms as they usually assume a light-tailed distribution of pixel
intensities (e.g., Gaussian distribution). A solution to this issue is the use
of a variance-stabilizing Anscombe transform
\cite{anscombe_transformation_1948} to make the noise variance uniform across
the whole image. For photon timestamp data, the variance-stabilizing transform
is the logarithm. See \nolink{\ref{sec:suppl_note_logT_estimator}} for a proof.
We design an image denoising deep neural network (DNN) that operates on
log-transformed first-photon timestamp images.

\begin{figure}[!htb]
    \centering
    \includegraphics[width=0.5\linewidth]{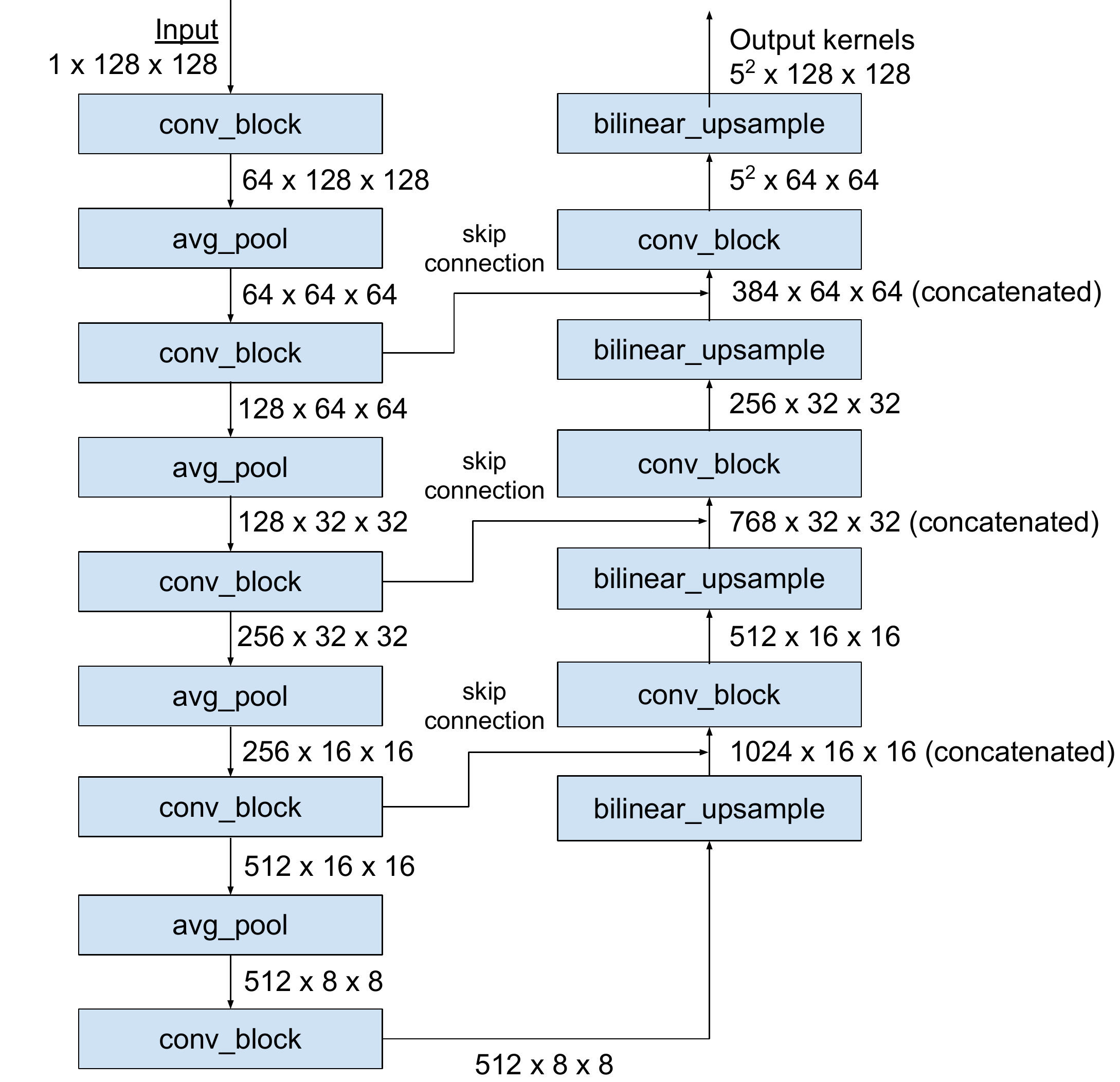}
    \caption{The kernel prediction network (KPN) architecture we have used to
      estimate per-pixel kernels of size $5 \times 5$, which is adapted from
      the architectures used for burst denoising in \cite{burstkpn_2018} and
      for denoising Monte-Carlo renderings in \cite{kpn_2017}. The input image
    size is $128 \times 128$ when training the network, but any image size can
    be used at the inference stage.}
    \label{fig:kpn_logtimg}
\end{figure}

We use a kernel prediction network (KPN) architecture 
\cite{kpn_2017,burstkpn_2018}. Our network architecture is shown in
Suppl.~Fig.~\nolink{\ref{fig:kpn_logtimg}}. The network produces $5 \times 5$
kernels for every pixel in the input image, which we apply to generate the
denoised image. The only substantial post-processing step is to correct the
bias introduced by using the log-timestamp instead of the timestamp itself (see
\nolink{\ref{sec:suppl_note_logT_estimator}}).

We train the network with timestamp images simulated from the DIV2K dataset
\cite{DIV2K_Intro,DIV2K_Report}. This dataset has 800 high-resolution images;
we simulate four random timestamp images for each image in the dataset for a
total of 3200 training images. The original 8-bit images are first converted to
16-bit linear luminance \cite{imagemagick}, before simulating the timestamps.
The simulated timestamps are then log-transformed.

We use the Adam optimizer \cite{kingma2017adam} with a learning rate of
$10^{-4}$. The loss function is a sum of squared errors in the pixel
intensities and absolute errors in the pixel-wise image gradients, both with
respect to the original image from which the timestamps are simulated
\cite{burstkpn_2018}.  Training runs for 1920 iterations with a batch size of 5
images, for a total of 3 epochs. Images are randomly cropped into $128 \times
128$ patches before passing into the network when training. However, since
the network is fully convolutional, it can handle arbitrary input image sizes
at test time.

The architecture of our kernel prediction network-based denoising DNN is
shown in Suppl. Fig.~\ref{fig:kpn_logtimg}.

\begin{figure}[!htb]
  \centering
  \includegraphics[width=0.80\linewidth]{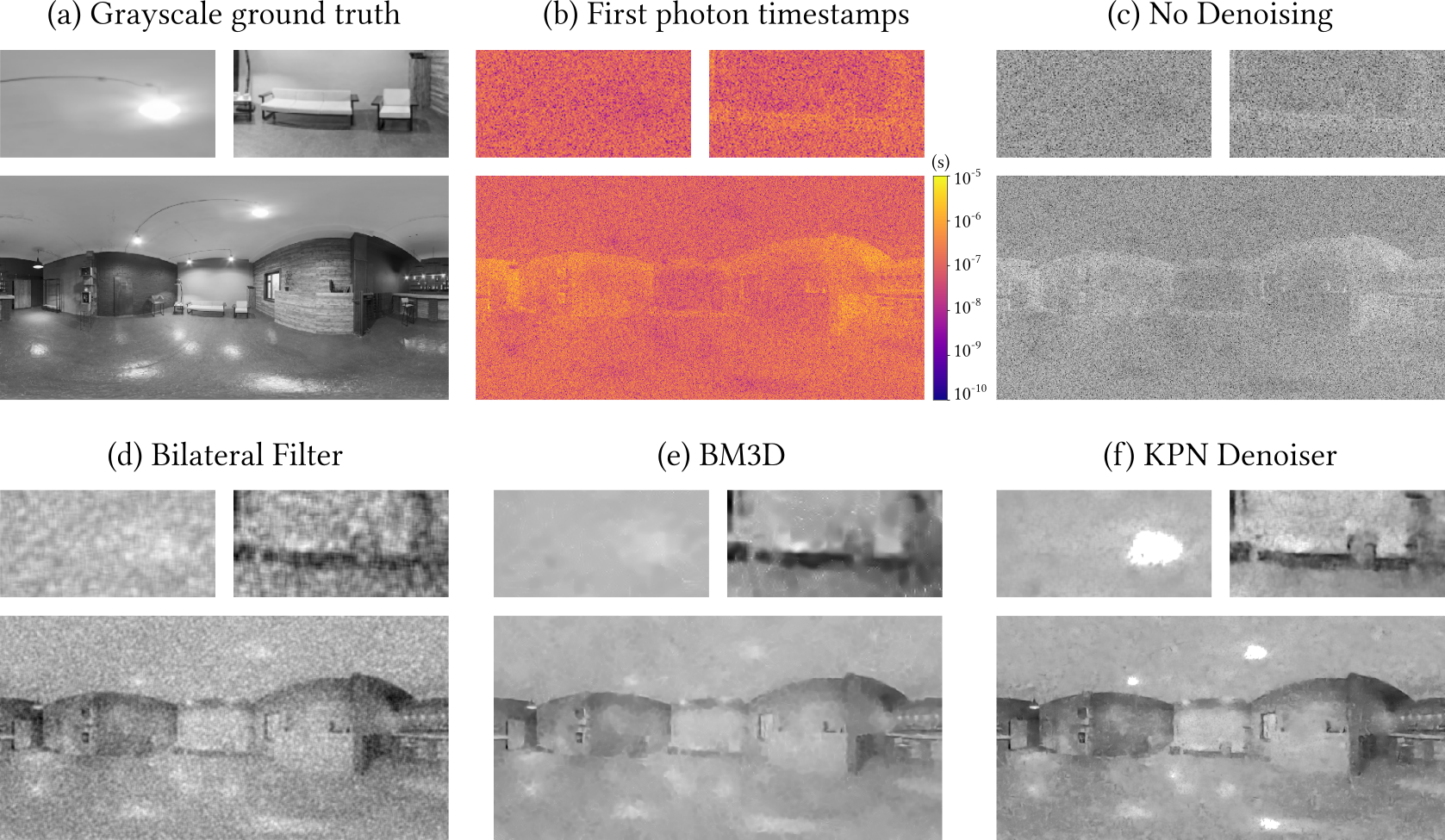}
  \caption{\textbf{Denoising IP-SPAD Images with Low Photon Counts:} (a-b) We simulate
  the extreme case of IP-SPAD imaging by sampling at most one photon
  time stamp per pixel of a ground truth image. (c) Simply inverting the each
  time stamp is not enough due to extreme noise, (d-f) so it is necessary to
  combine time stamps spatially. (d) We apply a bilateral filter ($\sigma=7$),
  which incorporates some spatial information, but still remains quite noisy.
  (e) BM3D \cite{Kostadin2006} may over smooth, and it seems to have particular
  trouble in bright regions. (f) Our KPN denoiser trained on photon timestamp
  data preserves some object shapes like the bright ceiling light and the
  couches.
  \label{fig:passive_first_photon}}
\end{figure}

Suppl. Fig.~\ref{fig:passive_first_photon} shows simulated denoising results
comparing our KPN-based denoiser with two standard denoising methods: bilateral
filtering and BM3D.

\clearpage
\subsection{Homoskedasticity of log-timestamps\label{sec:suppl_note_logT_estimator}}
In this section we show that the log-transformation is a variance stabilizing
transformation for first-photon timestamp data. Let $Y\sim\text{Exp}(\Phi)$ be
the arrival time of the first incident photon (we drop the subscript in $Y_1$
for simplicity). Our goal is to show that the variance of $\log(Y)$ is constant
(homoskedasticity).

\begin{align}
  \intertext{From the properties of the exponential distribution,}
    P(Y \leq t) = & 1 - e^{-\Phi t} \\
    \implies P(\log Y \leq \log t) = & 1 - e^{-\Phi t}. \\
  \intertext{Defining $\tilde Y = \log(Y)$ and $y = \log(t)$,}
    P(\tilde Y \leq y) = & 1 - e^{-\Phi e^y} \\
    \implies p_{\tilde Y}(y) = & \Phi e^{-\Phi e^y} e^y \\
  \intertext{where $p_{\tilde Y}(y)$ is the p.d.f. of $\tilde Y$.}
    \implies E[\tilde Y] =& \int_{-\infty}^{\infty} y \Phi e^{-\Phi e^y} e^y \, dy \\
                =& \int_{-\infty}^{\infty} \Phi y e^y e^{-\Phi e^y}\, dy. \\
    \intertext{Take $e^y = u$}
    E[\tilde Y] =& \int_{0}^{\infty} \Phi \log(u) e^{-\Phi u}\, du \\
    \intertext{Take $\Phi u = v$}
    E[\tilde Y] =& \int_{0}^{\infty} (\log(v) - \log(\Phi)) e^{-v} dv \\
        =& -\log(\Phi) + \int_{0}^{\infty} \log(v) e^{-v} dv \\
        =& -\log(\Phi) - \gamma, \\
  \intertext{where the second expression is an integral known to evaluate to
    $-\gamma$ ($\gamma \approx 0.577$ is the Euler-Mascheroni constant). We can
    see that the log-timestamp only has a constant bias away from the true
    log-timestamp ($= \log(\nicefrac{1}{\Phi})$), which can be removed
    separately.}
    \intertext{We repeat the same ideas for calculating the variance:}
    E[\tilde Y^2] =& \int_{-\infty}^{\infty} y^2 \Phi e^{-\Phi e^y} e^{y}\, dy \\
    \intertext{Take $e^y = u$ again:}
    E[\tilde Y^2] =& \int_{0}^{\infty} \Phi (\log^2 u) e^{-\Phi u}\, du \\
    \intertext{and $v = \Phi u \implies \log u = \log v - \log \Phi$. Then}
    E[\tilde Y^2] = & \int_{0}^{\infty} \log^2(v) e^{-v}\, dv \\
                & - 2 \log(\Phi) \int_{0}^{\infty} \log(v) e^{-v}\, dv \nonumber\\
                & + \log^2(\Phi) \nonumber \\
        = & \gamma^2 + \nicefrac{\pi^2}{6} + 2 \gamma \log(\Phi) + \log^2(\Phi), \\
    \intertext{where the first term on the right-hand side is also a known
    standard integral. Finally we have}
    \mathrm{Var}(\tilde Y) = & E[\tilde Y^2] - E[\tilde Y]^2 = \nicefrac{\pi^2}{6},
\end{align}
which proves the homoskedasticity of $\tilde Y = \log(Y)$.


\clearpage
\section{Hardware Prototype\label{suppl:hardware}}
\begin{figure}[!ht]
  \centering \includegraphics[width=0.4\columnwidth]{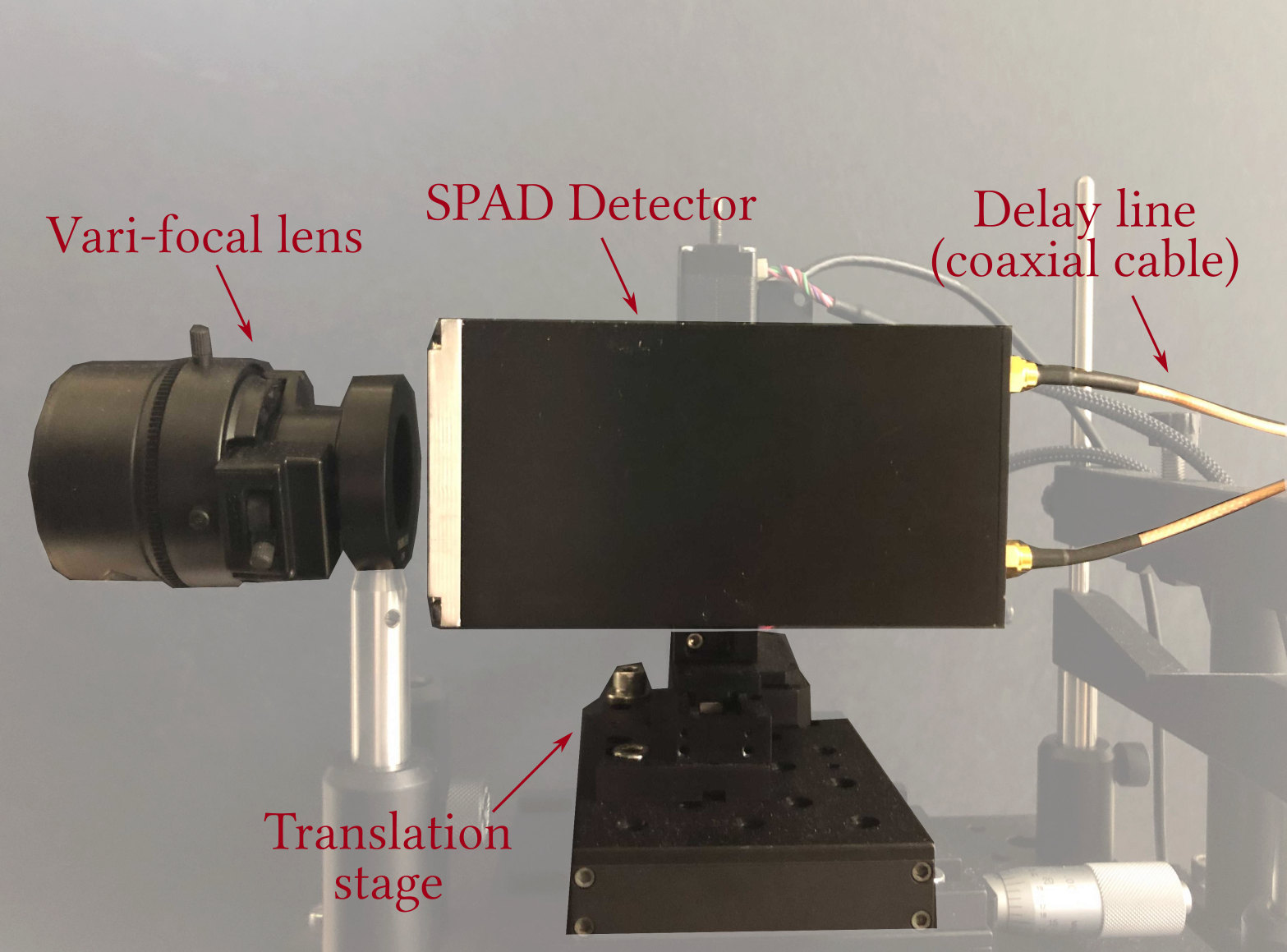}
  \caption{The IP-SPAD hardware prototype consists of a single SPAD pixel mounted on
    translation stages to scan the image plane of a vari-focal lens
    (Fujinon DV3.4x3.8SA-1). Part of a $\SI{20}{\m}$ long co-axial cable used for
    generating the dead-time delay is also shown.
    \label{fig:hardware_photo}}
\end{figure}

Our hardware prototype (\ref{fig:hardware_photo})
consists of a single SPAD pixel mounted on two
translation stages. Dead-time is controlled using a long cable that produces
analog delay.  After each photon detection event the SPAD pixel has to be kept
disabled for few tens of \si{\ns} to lower the probability of afterpulses
\cite{cova1991} and reset its original bias condition. Usually this dead-time
is set either using the discharge time of an R-C network or employing a digital
timing circuit, since the dead-time accuracy is not a limiting factor in
conventional SPAD applications. In case of dead-time defined using digital
timing circuits, there are implementations where its accuracy depends on the
period of an uncorrelated (with respect to photon arrival times) digital clock.
For example, a \num{100}~\si{\mega\hertz} clock frequency will limit the
accuracy of the dead-time to about \num{10}~\si{\ns}, which is too coarse to
get reliable photon flux estimates.  This is true especially at extremely high
photon flux values where photons get detected almost immediately after each
dead-time duration ends. As described in the main text, we rely on low-jitter
voltage comparators and analog delays introduced by long coaxial cables to
obtain precisely controlled dead-time durations with low jitter.

\section{Pixel Non-idealities}
When conducting experiments with our hardware prototype we found two
non-idealities: dead-time drift and non-zero gate rise time.  
\subsection*{Dead-time Drift}\label{suppl:dead time drift}
When imaging high flux regions for extended periods of time our hardware
prototype's dead-time increases; we call this \textit{dead-time drift}.  This
is due to heating of the SPAD front-end. We calibrated each pixel position
individually by constructing an inter-photon timing histogram and using the
first non-zero bin of this histogram as an estimate of the true dead-time for
that pixel position. Experimentally, we observed that the dead-time drift is
slower than the 5ms exposure times used so this method should approximate the
true dead-time well for each pixel position. Without this correction the error
introduced by the drift dominates the denominator in
Eq.~(\ref{eq:flux_estimator}) at high flux values, limiting the dynamic range.

\begin{figure}[!ht]
  \centering \includegraphics[width=0.6\columnwidth]{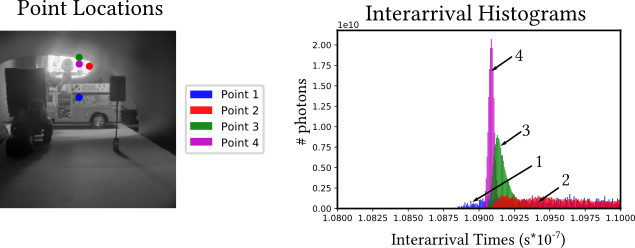}
  \caption{This figure shows inter-photon histograms for 4 points from the
    tunnel scene.  Notice that the histograms are not aligned on the left edge,
    indicating a drift in dead-time. We correct for this drift by taking the
    time of the first non-zero bin as the dead-time for that pixel.
    \label{suppl_fig:drift_rise}}
\end{figure}

When our single-pixel IP-SPAD stays active for long periods of time dead-time
drift becomes a problem. Suppl. Fig.~\ref{suppl_fig:drift_rise} shows
inter-photon timing histograms of four different scene points with increasing
flux levels ($1\rightarrow 4$). Notice that the histograms are not aligned on
the left edges indicating a difference in dead-times at these points. We
correct for the dead-time drift in by using a dead-time estimate for each pixel
in the final image. We set the dead-time estimate in a pixel to the smallest
inter-arrival time from that pixel, this has the effect of shifting each
pixel's inter-arrival histogram to zero.  In the tunnel scene the difference
between the longest used and shortest used dead-time is \SI{887}{\pico\second},
a variation of about 0.8\%.

\subsection*{Gate Rise Time}\label{suppl:rise time}
When the SPAD enters and exits the dead-time phase, its bias voltage
has to be quickly changed from above to below the breakdown value, and
vice-versa \cite{Buttafava2014}. The duration of these transitions is as
critical as the dead-time duration itself, and has to be short (in order to
swiftly restore the SPAD bias for the next detection) and precise (to prevent
variations in the overall dead-time duration). In our system the rise
times are on the order of \SI{200}{\pico\second}:
it translates into non-exponentially shaped inter-photon timing histograms,
especially in high flux regions. We did not find that this behavior
detrimentally effected our results; however, it has an effect similar to
slightly tone mapping bright regions downward. 

Unlike dead-time drift, the rise time behavior seems to be independent of how
long the SPAD was exposed to a high flux source. Fig.  \ref{fig:exp_hists}
shows inter-photon timestamp histograms for increasing photon flux levels. Rise
time causes these to deviate from an exponential shape at high flux levels. 

We found that this behavior made it virtually impossible to fully saturate the
SPAD pixel, that is increasing the incident flux would lead to a non-linear
increase in photons counted. We performed an experiment where a laser was
directly pointed into the SPAD active region and the power of the laser was
increased. We found that the photon counts did not saturate before the SPAD
overheated and shut itself off. 

The rise-time behaviour can by incorporated into the flux estimator derived in
\ref{suppl:MLE_conditional_derivation} using a time-varying quantum efficiency
$q(t)$. For $t<0$, $q(t)=0$ and  $\int_0^{\infty} q(t) dt \rightarrow\infty$.
When the dead time ends, the IP-SPAD pixel's $q(t)$ ramps up to its peak value.
The probability distribution of time-of-darkness, $Y_i$, can be written as:

\begin{equation}
Y_{i} \sim f_{Y_{i}}(Y_i | Y_{1}\dots Y_{i-1})=\begin{cases}
q(Y_i)\Phi e^{-\Phi\int_0^{Y_i}q(l)dl} & \mbox{for } 0< Y_i< T_i\\
e^{-\Phi \int_0^{T_i}q(l)dl} \,\cdot\, \delta(t-T_i) &\mbox{for } Y_i=T_i\\
0 & \mbox{otherwise.}
\end{cases} \label{eq:interarrival_distr_varyQ}
\end{equation}
where $T_i$ is defined in \ref{suppl:MLE_conditional_derivation}. For a series
of $N$ timestamps with times-of-darkness given by $Y_1\dots Y_{N}$, we use a
similar derivation to \ref{suppl:MLE_conditional_derivation} to find the
maximum likelihood estimator (MLE): 
\begin{equation}
  \widehat \Phi =  \frac{N}{\int^{T_{N+1}}_0 q(t)dt+\sum_{i=1}^N\int^{Y_i}_0 q(t)dt}. \label{suppl:MLE_varyQ}
\end{equation}
Eq.~(\ref{suppl:MLE_varyQ}) reduces to Eq.~(\ref{eq:MLE_cond}) if $q(t)$ is an
ideal step function. For the experimental results shown in the main text, the
IP-SPAD pixel's $q(t)$ was not precisely known so we could not apply this
correction. Future work will look at estimating $q(t)$ from inter-photon
histograms and quantifying SNR improvements from such a correction.

\clearpage
\section{Additional Results\label{suppl:additional_results}}

\begin{figure}[!htb]
  \centering
  \includegraphics[width=0.55\linewidth]{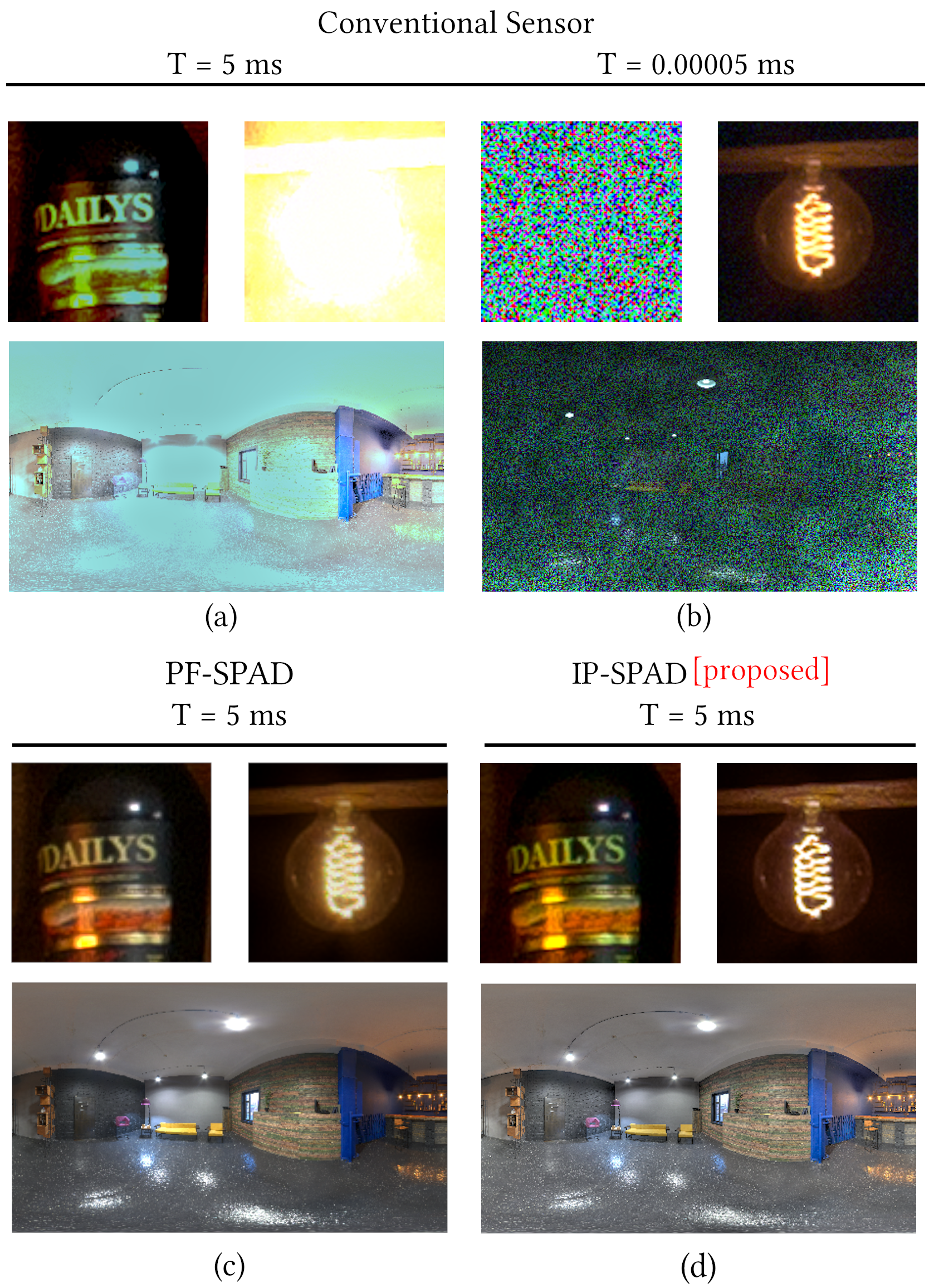}
  \caption{\textbf{Simulated Extreme Dynamic Range Scene:}
  This figure shows simulated extreme dynamic range images using an IP-SPAD
  camera compared with a conventional camera with different exposure settings.
  (a) A \SI{5}{\ms} exposure image with a conventional camera (full-well capacity
  34,000 and read noise 5$e^-$ has many saturated pixels. Observe that the
  bright bulb region is washed out.  (b) A short exposure image is dominated by
  shot noise in darker parts of the scene.  It becomes visible only at a much
  lower exposure setting. Since this is a simulation we were able to reduce the
  exposure time down to $5 \times 10^{-5}$ \si{\ms} which may be impossible to
  achieve with a conventional camera. In practice, this exposure can be
  achieved by, say, reducing the shutter speed to 1/16,000 and adding a 10-stop
  ND filter.  (c) A PF-SPAD camera is able to capture both dark and bright
  regions in a single exposure, but the bright bulb filament still suffers from
  noise due to the soft-saturation phenomenon. (d) Our proposed IP-SPAD method
  estimates scene brightness using high-resolution timestamps to capture both
  extremely dark and extremely bright pixels, beyond the soft-saturation limit
  of a counts-based PF-SPAD.  (Original image from HDRIHaven.com) 
    \label{fig:sim_hdr_suppl}}
\end{figure}
\clearpage

\begin{figure}[!htb]
  \centering
  \includegraphics[width=0.95\linewidth]{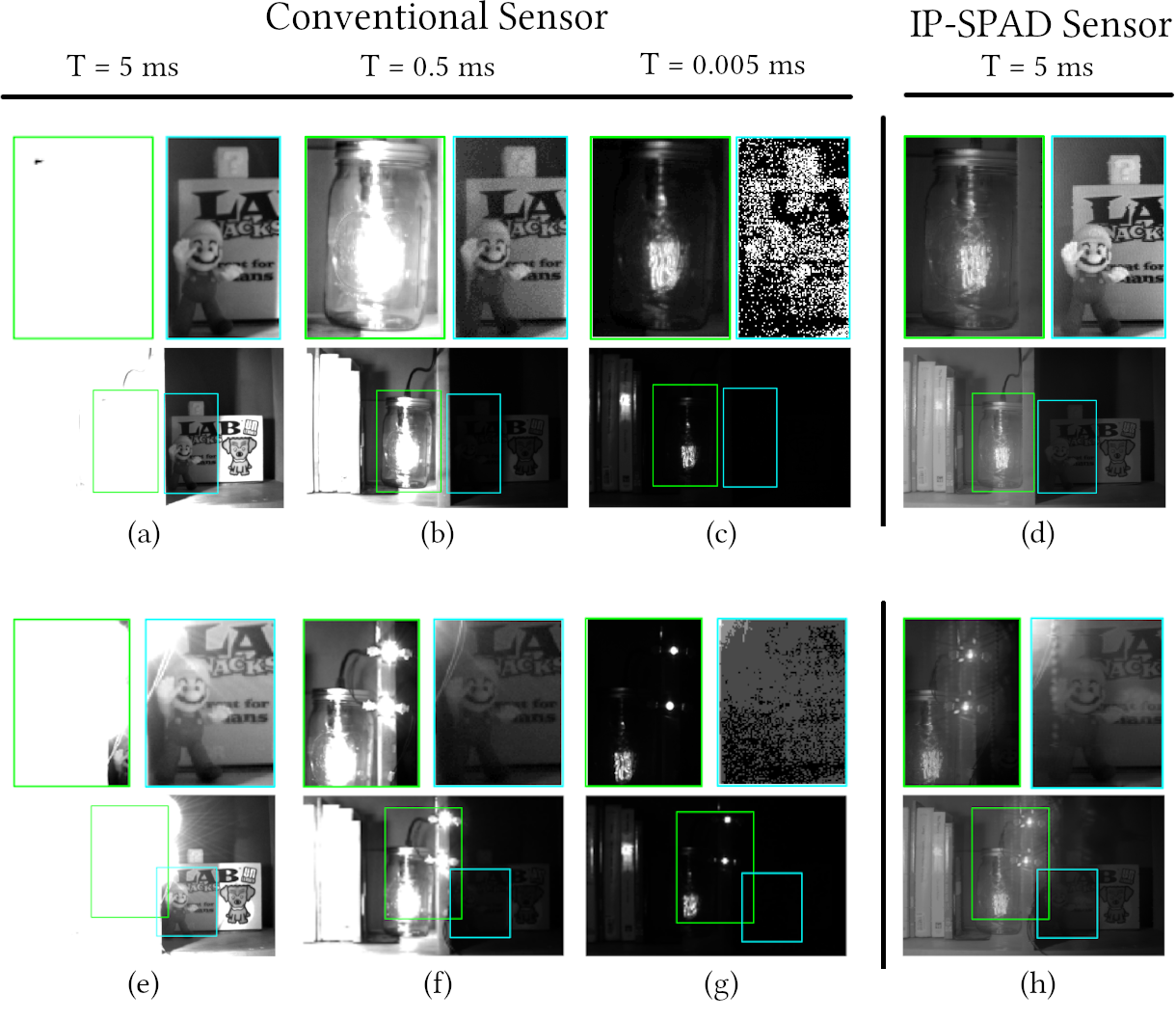}
  \caption{\textbf{Experimental Extreme Dynamic Range ``Shelf'' Scene:}
  This ``Shelf'' scene shows extreme dynamic range, with a bright bulb filament
  in one of the shelves and text in the neighboring shelf which is dark and not
  directly illuminated by the light source. The bottom row of images uses a
  similar setup as the top row but also includes two bright LED lights in
  addition to the filament bulb.  The conventional camera requires three
  exposures to cover the dynamic range of this scene. The proposed IP-SPAD flux
  estimator captures the scene in a single exposure.
    \label{fig:shelf_expt}}
\end{figure}

\clearpage

\begin{figure}[!htb]
  \centering
  \includegraphics[width=0.8\linewidth]{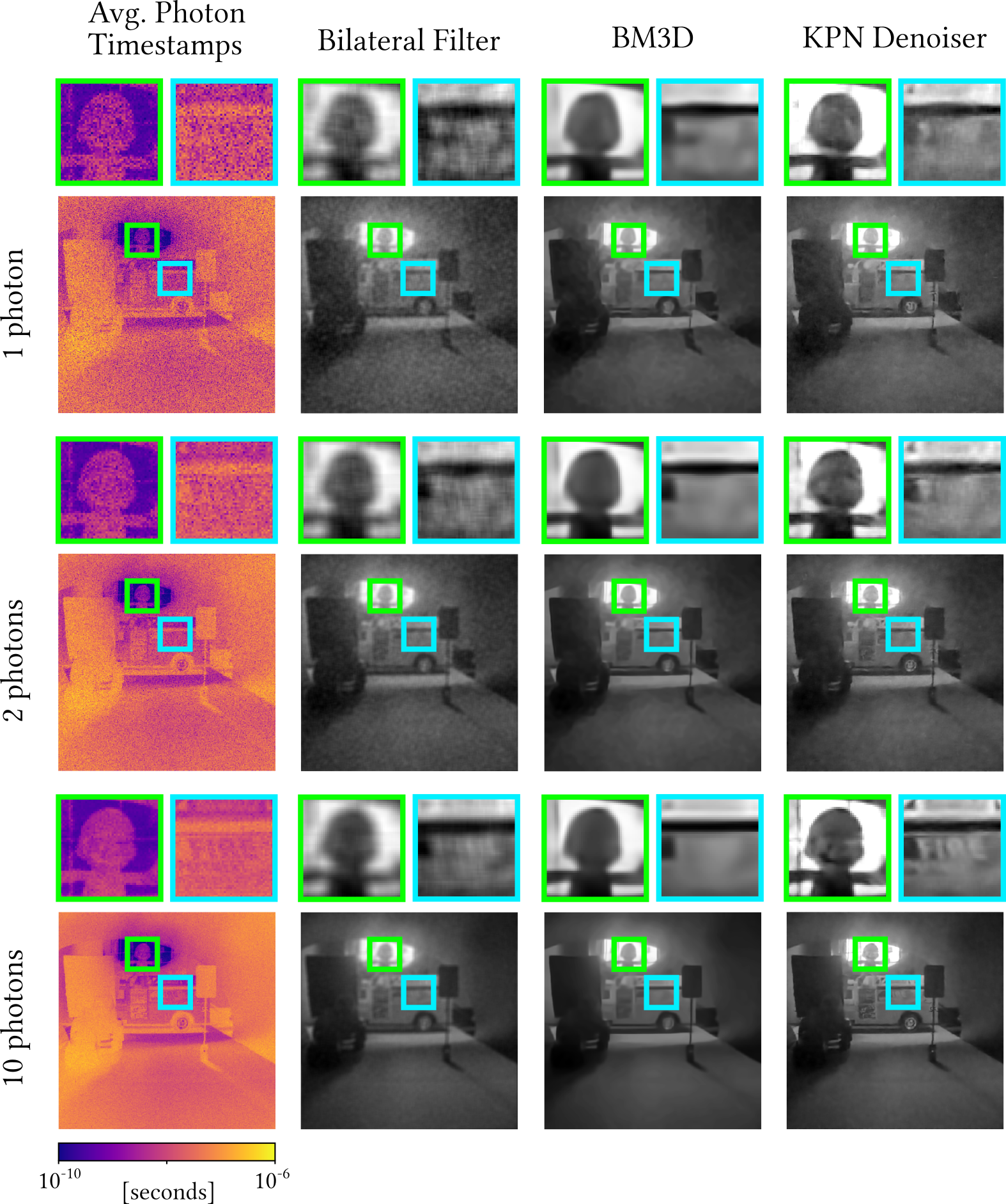}
	\caption{\textbf{Effect of Increasing Number of Photons on Denoising.}
  Some image details start appearing with as few as 10 photon timestamps per
  pixel. For example, the text on the fire-truck is visible with images
  denoised with the bilateral filter and our KPN-based denoiser.  BM3D appears
  to give less noisy results in this example but finer details are lost.
  \label{fig:passive_first_photon_expt_full}}
\end{figure}

\clearpage
\section{Pixel Designs for Passive SPAD Imaging\label{suppl:pixel_designs}}

\begin{figure}[!htb]
  \centering \includegraphics[width=0.61\columnwidth]{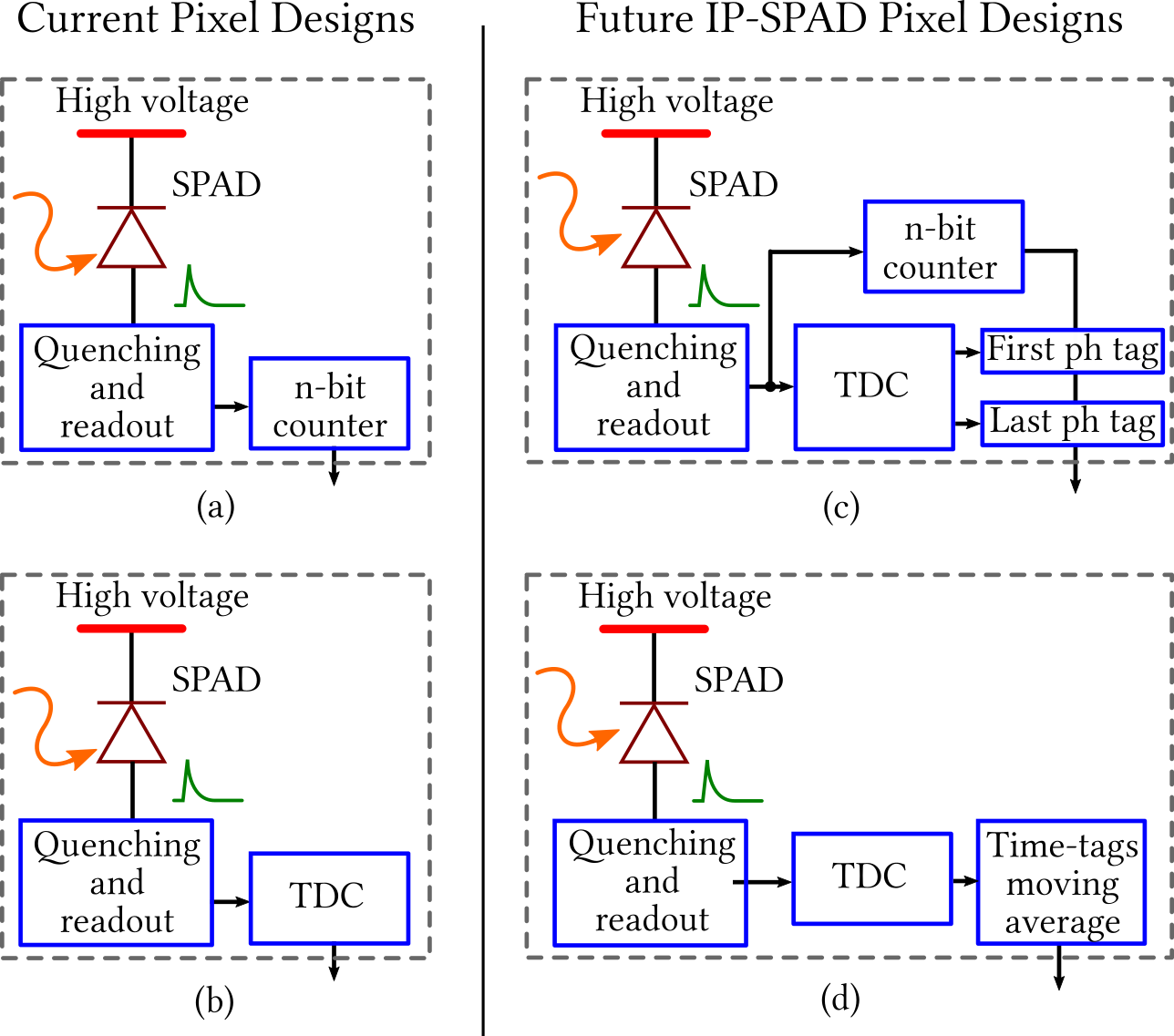}
  \caption{\textbf{IP-SPAD pixel designs for passive imaging:} (a) and (b) are
  existing SPAD pixel designs with counts and in-pixel timing circuits. (c) and
  (d) are hypothetical future pixel designs for passive IP-SPAD cameras that
  store individual photon timestamps or compute summary statistics on the fly.
  \label{fig:array_designs_full}}
\end{figure}

\smallskip
\noindent\textbf{Passive SPAD Pixel Architectures}
Many current SPAD pixel designs are targeted towards specialized active imaging
applications that operate the detector in synchronization with a light source,
such as pulsed laser. The most common data processing task is to generate a
\emph{timing histogram} which counts the number of photons detected by the SPAD
pixel as a function of the (discretized) time delay since the transmission of
the most recent laser pulse. The requirements for the passive imaging technique
shown in this paper are different: there is no pulsed light source to provide
a timing reference. Instead, it is important to precisely control (1) the dead
time duration (2) rise and fall times of the SPAD bias circuitry, and
(3) the duration of the global exposure time.

Our single-pixel IP-SPAD hardware prototype, although acceptable as a
proof-of-concept, is not a scalable solution for sensor arrays.
Delay-locked loops circuits suitable for multi-pixel implementation
can be used in the future to precisely control the dead-time duration.  A large
array of IP-SPAD pixels will generate an unreasonably large volume of raw
photon timestamp data that cannot be transferred off the sensor chip for
post-processing. A megapixel SPAD array has been recently demonstrated using a
\SI{180}{\nm} CMOS technology \cite{Morimoto_2020}, but the in-pixel
electronics is currently limited to gating circuitry and a 1-bit data register.
The trade-off between SPAD performance and pixel number can be overcome by
recently-developed 3D-stacking approaches where  SPAD arrays are fabricated in
a dedicated technology, the high density data-processing electronics are
developed in scaled technology, and then the two chips/wafers are mounted one
on top of the other \cite{Henderson_2019_ISSCC,Charbon_2018}.

Fig.~\ref{fig:array_designs_full}(a) shows the simplest single-pixel architecture
currently used as a building-block in large SPAD arrays. It comprises the
photodetector, its readout and quenching circuits and a digital counter, for
storing the number of detected photons. While this architecture is widely used
\cite{bronzi2015spad}, it does not exploit photon arrival times to increase
dynamic range. As shown in Fig.~\ref{fig:array_designs_full}(b), adding an
in-pixel time-to-digital converter (TDC) able to acquire and store individual
photon time-stamps (with respect to the exposure time synchronization signal)
can solve this limitation.  Also this approach is nowadays quite common when
designing SPAD arrays \cite{Henderson_2019, Portaluppi_2018}, however,
increasing the array dimension and considering a very high incident photon
flux, it will be impractical to acquire and transfer timestamps for each photon
and each pixel, because it will lead to intractable volume of data to be
processed. Instead, a more efficient way of storing and transmitting photon
time-stamp data for passive imaging can rely on simply storing the first and
last photon time-stamps within a single exposure time, together with the total
photon counts. The corresponding pixel design is shown in
Fig.~\ref{fig:array_designs_full}(c). While this increases pixel complexity
over the previous SPAD pixel design examples, it only requires two additional
data registers. The disadvantage of this scheme is that, depending on the total
exposure time, the TDC may require a large full-scale range.  For example,
using an exposure time in the millisecond range and the timestamp resolution in
picoseconds, the TDC data depth will be $\log_2(10^{-3}/10^{-12})\approx
30$~bits.

Note that our brightness estimator keeps track of the average time-of-darkness
between photon detections over a fixed exposure time. An alternative to storing
first and last timestamps may be to instead store a running average of the
inter-photon times, as shown in Fig.~\ref{fig:array_designs_full}(d). This can be
implemented in-pixel using basic digital signal processing circuits. At high
photon flux levels, the expected inter-photon times will be short enough that a
TDC with smaller full scale range could be used. Although the inter-photon
times may still be quite long for low flux levels, the flux estimator can fall
back to using photon counts only, instead of timestamps.

\paragraph{SPAD Array Designs for Passive Imaging}
The theoretical analysis and experimental results in this paper were restricted
to a single SPAD pixel. For most passive imaging applications, in practice,
there will be a need to scale this method to large form factor SPAD arrays with
thousands of pixels. This will introduce additional design challenges and noise
sources not discussed in this work. A large form-factor SPAD array of
free-running SPAD pixels will generate an unreasonably large volume of raw
photon time-stamp data that cannot be simply transferred off the sensor chip
for post-processing. For instance, consider a hypothetical 1 megapixel SPAD
array consisting of pixels shown in Fig.~\ref{fig:array_designs_full}(b), with dead
time of \num{100}~\si{\ns}. Assume an average photon flux of \num{e5} photons/s
over the pixel array and the pixels generate \num{32}-bit IEEE floating-point
timestamps for each detected photon. This corresponds to
\num{400}~\si{\gibi\byte\per\s} of data generated from the chip. A megapixel
SPAD array has been recently demonstrated using a 180nm CMOS technology
\cite{Morimoto_2020} , but the in-pixel electronics is currently limited to
gating circuitry and 1-bit data register (photon detected or not). 

One possible solution to overcome this problem could include the design of
large arrays using a combination of pixel architectures sketched in
Fig.~\ref{fig:array_designs_full}, i.e. where only a fraction of pixels would
include high resolution TDCs while the rest of the pixels only use photon
counters.  This will still enable capturing extremely high flux values albeit
with reduced spatial resolution. In another solution TDCs are shared among more
pixels, while counters are integrated in each pixel. This will reduce the
maximum count rate, but each detected photon is counted and time tagged.

SPAD performance (i.e. detection efficiency, dark count noise, temporal
resolution, afterpulsing probability) in developing multi-pixel arrays is
usually better when using ``legacy'' fabrication technologies, like 350 nm and
180 nm CMOS, or even ``custom'' technologies (which, however, do not allow the
on-chip integration of ancillary electronics) \cite{Ghioni_2007}. With such
technologies, the relatively large minimum feature size prevents in-pixel
integration of sophisticated electronics like high-resolution (few \si{\ps})
TDCs, data processing circuits and memories (unless without accepting an
extremely low fill-factor). The trade-off between SPAD performance and pixel
number can be overcome by recently-developed 3D-stacking approaches:  SPAD
array is fabricated in a dedicated technology, the high density data-processing
electronics is developed in scaled technology, and then the two chips/wafers
are mounted one on top of the other \cite{Henderson_2019_ISSCC,Charbon_2018}.

Passive IP-SPAD arrays may also require pixel-wise calibration. The non-linear
pixel response curve may make this more challenging than conventional CMOS
camera pixels. It will be necessary to characterize non-uniformity in terms of
dead-time durations and timing jitter and account for these for removing any
fixed pattern noise.

Another important practical consideration is power requirement, especially when
operating in high flux conditions where a large number of avalanches will be
created causing huge power requirement for processing these in real-time and
reading out the counts. There is also a significant heat dissipation issue
which can exacerbate pixel calibration due to the strong temperature dependence
of various pixel parameters like dark count rate and dead-time drifts. Such
power issues may be mitigated with scaled technologies operating at lower
supply voltage.

\end{document}